\documentclass{article}

\usepackage{microtype}
\usepackage{graphicx}
\usepackage{subcaption}
\usepackage{booktabs} 
\usepackage{multirow}

\usepackage{hyperref}
\usepackage{caption}
\usepackage{algorithm}
\usepackage{algpseudocode}

\algrenewcommand\algorithmicrequire{\textbf{Input:}}
\algrenewcommand\algorithmicensure{\textbf{Output:}}

\algnewcommand\Not{\textbf{not} }

\long\def\eat#1{}

\usepackage{listings}
\newcommand{\TODO}[1]{\textcolor{red}{#1}}
\newcommand{\CUT}[1]{\textcolor{yellow}{#1}}
\newcommand{\IGNORE}[1]{}

\newcommand{\ourname}[0]{\textsc{Tensat}}

\newdimen\figrasterwd
\figrasterwd\textwidth

\usepackage{amsmath}
\usepackage[capitalize]{cleveref}
\usepackage[accepted]{mlsys2021}

\mlsystitlerunning{Equality Saturation for Tensor Graph Superoptimization}
\begin{document}

\twocolumn[
\mlsystitle{Equality Saturation for Tensor Graph Superoptimization}




\mlsyssetsymbol{equal}{*}

\begin{mlsysauthorlist}
\mlsysauthor{Yichen Yang}{equal,mit}
\mlsysauthor{Phitchaya Mangpo Phothilimthana}{goo}
\mlsysauthor{Yisu Remy Wang}{uw}
\mlsysauthor{Max Willsey}{uw,octoml}
\mlsysauthor{Sudip Roy}{goo}
\mlsysauthor{Jacques Pienaar}{goo}
\end{mlsysauthorlist}

\mlsysaffiliation{mit}{MIT EECS \& CSAIL}
\mlsysaffiliation{goo}{Google, Mountain View, CA, USA}
\mlsysaffiliation{uw}{University of Washington, Seattle, USA}
\mlsysaffiliation{octoml}{OctoML, Seattle, USA}

\mlsyscorrespondingauthor{Yichen Yang}{yicheny@mit.edu}
\mlsyscorrespondingauthor{Phitchaya Mangpo Phothilimthana}{mangpo@google.com}
\mlsyscorrespondingauthor{Yisu Remy Wang}{remywang@cs.washington.edu}

\mlsyskeywords{Machine Learning, MLSys}

\vskip 0.3in

\begin{abstract}
One of the major optimizations employed in deep learning frameworks is graph rewriting.
Production frameworks rely on heuristics to decide if rewrite rules should be applied and in which order. 
Prior research has shown that one can discover more optimal tensor computation graphs if we search for a better sequence of substitutions instead of relying on heuristics.
However, we observe that existing approaches for tensor graph superoptimization both in production and research frameworks apply substitutions in a sequential manner.
Such sequential search methods are sensitive to the order in which the substitutions are applied and often only explore a small fragment of the exponential space of equivalent graphs.
This paper presents a novel technique for tensor graph superoptimization that employs \emph{equality saturation} to apply all possible substitutions at once. We show that our approach can find optimized graphs with up to 16\% speedup over state-of-the-art, while spending on average 48x less time optimizing.

\end{abstract}
]


\begin{NoHyper}
\printAffiliationsAndNotice{\textsuperscript{*} Work done during internship at Google }  
\end{NoHyper}

\section{Introduction}

Deep learning frameworks and compilers 
have enabled diverse kinds of machine learning models to run efficiently on numerous compute platforms. Neural network models in these frameworks are  typically represented as tensor computation graphs. To improve the runtime performance of a tensor graph, these frameworks perform various important optimizations, among which is graph rewriting.
Graph rewriting takes in a tensor graph $g$ and a set of semantics-preserving graph rewrites $R$.
By applying rewrites to $g$, it seeks to find a semantically equivalent $g'$ with a lower cost according to some objectives.
The current industry-standard approach adopted by most frameworks is to use a manually curated set of rewrite rules and rely on a heuristic strategy to determine the order in which to apply the rewrite rules.
However, this approach often leads to sub-optimal results both due to the non-comprehensive set of rewrite rules, as well as the sub-optimal graph substitution heuristic \cite{taso,metaflow}. 

This paper aims to address the sub-optimality problem of graph rewrite strategies, while leveraging the existing rewrite rules generation technique \cite{taso}.
Prior research has shown that searching for sequences of substitutions
\cite{taso,metaflow,Fang:sampling} outperforms heuristic approaches.
However, both heuristic- and search-based solutions rely on sequential application of substitutions.
Since rewrites often depend on or enable one another, 
optimization depends heavily on the order in which rewrites are applied;
this classically tricky problem is known in the compilers community as the ``phase-ordering'' or ``rewrite-ordering'' problem.

\begin{table}[]
    \centering
    \footnotesize
    \begin{tabular}{ccc|cc}
    \toprule
    & \multicolumn{2}{c|}{\bf Search time (s)} & \multicolumn{2}{c}{\bf Runtime speedup (\%)} \\
    & TASO & \ourname{} & TASO & \ourname{} \\
    \midrule
    NasRNN & 177.3 & \textbf{0.5} & 45.4 & \textbf{68.9} \\
    BERT & 13.6 & \textbf{1.4} & 8.5 & \textbf{9.2} \\
    ResNeXt-50 & 25.3 & \textbf{0.7} & 5.5 & \textbf{8.8} \\
    NasNet-A & 1226 & \textbf{10.6} & 1.9 & \textbf{7.3} \\
    SqueezeNet & 16.4 & \textbf{0.3} & 6.7 & \textbf{24.5} \\
    VGG-19 & 8.9 & \textbf{0.4} & \textbf{8.9} & \textbf{8.9} \\
    Inception-v3 & 68.6 & \textbf{5.1} & 6.3 & \textbf{10.0} \\
    \bottomrule
    \end{tabular}
    \caption{Comparison of optimization time and runtime speedup of the optimized computation graphs over the original graphs, TASO~\cite{taso} v.s. \ourname{}.}
    \label{table:ngraph}
\end{table}

This paper presents \ourname{}, a tensor graph superoptimization framework that employs \emph{equality saturation} \cite{eqsat, eqsat-llvm, egg},
a recent technique that mitigates the phase-ordering problem by
applying all possible rewrites at once.  
Equality saturation splits program optimization into two phases: {\em exploration} and {\em extraction}.
The exploration phase uses a data structure called \emph{e-graph} to compactly generate and store all rewritings of the input program.
The exploration can continue until \emph{saturation}, 
where the e-graph stores all possible ways to write the input program using a given set of rewrites.
Finally, the extraction phase selects from the e-graph the equivalent program with the lowest cost according to a given cost model. The compact representation of the exponentially large search space using e-graphs enables extraction algorithms to find the globally optimal equivalent program quickly.


Applying equality saturation to tensor graph rewriting requires non-trivial extensions in both the exploration and extraction phases.
We extend the exploration phase to support complex, non-local rewrite rules that are necessary to produce highly efficient tensor graphs. 
Additionally, we introduce a novel method to filter out invalid subgraphs from an e-graph,
which enables our extraction procedure based on Integer Linear Programming (ILP) to quickly find the optimal solution. 

We evaluated \ourname{} on a number of well-known machine learning models executing on a GPU\footnote{Code is available at: \url{https://github.com/uwplse/tensat}}.
As highlighted in Table~\ref{table:ngraph}, \ourname{} can synthesize optimized graphs that are up to 16\% faster in runtime than state-of-the-art \cite{taso} (up to 68.9\% faster than the unoptimized graph), while reducing the optimization time by up to 300x. 
By having the e-graph compactly representing an exponential number of equivalent graphs, \ourname{} is able to cover a larger search space more efficiently than the sequential search methods.
As a result, our search approach is both extremely effective and fast enough to be used as part of a normal compilation flow.



\eat {

\paragraph{Contributions} We summarize the key contributions 

\begin{itemize}
    \item We introduce a novel approach for tensor graph superoptimization based on equality saturation.
    \item We present several techniques to improve the efficiency of our equality saturation based approach. \TODO{mangpo: let expand on what the several techniques are. We probably want to highlight the multi-pattern support for equality saturation since previous work doesn't do that (we should double check on that), and it is not just about efficiency.}
    \item We evaluate our approach by comparing with the state of the art \cite{taso} on xxx benchmark models. We show that our approach is able to both discover better optimized graphs and use less time in doing so.
    \TODO{jp: What about: We implement our techniques in a graph optimization framework (\ourname) and show it improves on state-of-the-art optimization while requiring significantly less time to do so (up to 3.9x faster performing graphs found in up to 379x while exploring exponentially larger search space.}
\end{itemize}

Put the key results (teaser) upfront.

}

\eat {

    \subsection{Tensor Graph Superoptimization (describe in intro)}
    
    \begin{itemize}
        \item Describe problem formulation: original tensor computation graph, set of rewrite rules, a cost model (cost for each op), an evaluation mechanism for measuring full graph runtime.
        \item Goal: search for an optimized graph via applying the rewrite rules that has an improved full graph runtime. 
        \item Note that we do not consider the rule discovery part. We focus on the search part. 
        \item \TODO{mangpo: we should explain the problems of the current approach here or in the intro.}
    \end{itemize}

}

\eat{
\TODO{
Hyeontake's high-level comments:
(1) it was a little unclear whether the paper proposes equality saturation or it makes equality saturation practical & efficient for ML graph optimization and (2) it requires some more reading until I know how important is being able to search equiv. graphs in an online fashion.
I think both are presentation issues -- as you mention in the comment, highlighting your technical contributions and removing redundancy should help (1), and the intro might mention the end-to-end numbers and use the number of equivalent graphs considered to explain why it is possible.
}

\TODO{
Mangpo's high-level comments:
\begin{itemize}
    \item We should highlight our technical contributions (efficient cycle filtering and multi-pattern support) more in the technical sections. 
    \item (Cycle filtering) We should be clear that some of the prior eq-sat work uses ILP (Ross et al.). Our formulation is similar to their. They also use ILP to get rid of cycles.
    \item (Cycle filtering) The new part is that we introduce the efficient cycle filtering algorithm, so we don't need ILP to filter cycles, which is very important for scalability. Our algorihm should be applicable to eq-sat in general, not just for our problem.
    \item (Multi-pattern) What can we say about prior work? Do they use multi-pattern rules? How do they handle them? If we're the first to support multi-pattern rules, we should highligh that.
    \item There are some redundancy between the eq-sat background and eq-sat for Tenset.
    \item Make sure our terminology is consistent: rewrite rule = rewrite = pattern (with placeholder variables), substitution (instance of matched rewrite).
\end{itemize}
}
}

\section{Equality Saturation Background}\label{sec:esat}

Term rewriting is a time-tested approach to program optimizations.
\ourname{} employs \textit{equality saturation},
a recent technique that provides a search-based approach 
to term rewriting that can avoid problems encountered by the traditional approach.

\subsection{Term Rewriting}
In the term rewriting paradigm, the input to the optimizer is an initial program expression (term) $e$
and a set of rewrite rules $R$.
Each \emph{rewrite} in $R$ takes the form $l \to r$, where $l$ and $r$ are both \emph{patterns} -- terms from the program grammar with placeholder variables.
Each rewrite rule states the semantic equivalence between $l$ and $r$.
To apply a rewrite $l \to r$ to $e$, the optimizer searches for the pattern $l$ in $e$, 
yielding a list of \emph{matches}, where a 
match $\sigma$ maps variables in $l$ to subterms of $e$.
The matches can then be applied to the right-hand side, denoted $r[\sigma]$, to produce the subterms to be replaced in $e$.

\paragraph{Example} Let $e = (a \times 2) / 2$ and the rewrite $l \to r$ be $x \times 2 \to x \ll 1$.
To apply $l \to r$ to $e$, the optimizer first searches for $l$ in $e$, 
yielding a single match $\sigma = \{x \mapsto a\}$.
Then, the term $r[\sigma] = a \ll 1$ replaces the matched subterm in $e$,
giving the final result: $(a \ll 1) / 2$.

As the above example suggests, term rewriting for optimization suffers from the problem of choice: 
applying the wrong rewrite at the wrong time can ``hide'' optimizations from other rewrites.
In our example, the classical strength reduction $x \times 2 \to x \ll 1$ is beneficial in some contexts,
but in this case will prevent the ultimate goal of rewriting $(a \times 2) / 2$ to $a$.
The problem stems from the fact the term rewriting is typically destructive; 
i.e., applying a rewrite ``forgets'' the initial term.

\subsection{E-graphs}

An \textit{e-graph} is a data structure originally devised for use in theorem provers 
\cite{nelson, simplify}
that compactly encodes an equivalence relation over many terms.
An e-graph is a set of equivalence classes (\textit{e-classes}), each of which is a set of equivalent \textit{e-nodes}.
An e-node is an operator from the language paired with a list of children e-classes. 
An e-graph 
\textit{represents} the terms that can be seen by choosing a representative e-node for each e-class.
More formally, 
\begin{itemize}
\itemsep0em 
    \item An e-graph represents a term if any of its e-classes do.
    \item An e-class represents a term if any of its equivalent e-nodes do.
    All terms represented by an e-class are equivalent.
    \item An e-node $f(c_1, ..., c_n)$ represents a term $f(t_1, ..., t_n)$ if each e-class $c_i$ represents $t_i$. 
    A childless e-node $a$ represents a constant term $a$.
\end{itemize}

\begin{figure}
    \centering
    \includegraphics[width=0.85\linewidth]{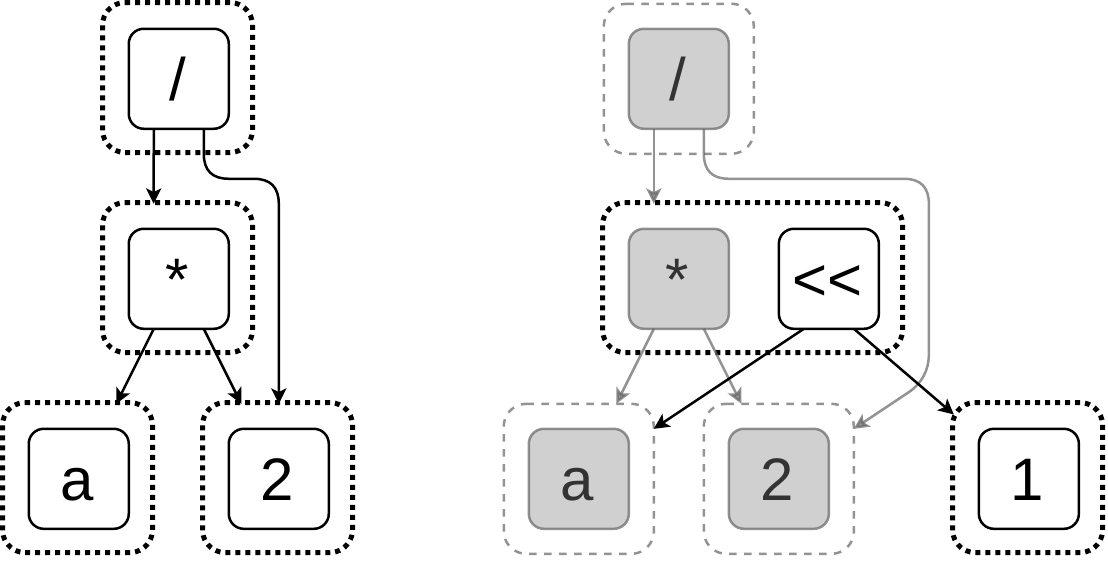}
    \caption{
      \textit{Left:} An e-graph representing the term $(a \times 2) / 2$.
      Dotted boxes show e-classes, and arrows connect e-nodes to their e-class children.
      \textit{Right:} The e-graph after applying the rewrite $x \times 2 \to x \ll 1$.
      Only a few e-nodes were added (highlighted in white), 
      and the result represents both the initial and rewritten terms.
    }
    \vspace{-0.4em}
    \label{fig:egraphs}
\end{figure}

\autoref{fig:egraphs} shows an e-graph that represents the term $(a \times 2) / 2$.
The e-class containing the division e-node is called the \textit{root} e-class since it represents our initial term. 
\autoref{fig:egraphs} also demonstrates how e-graphs can support rewriting.
Similar to traditional term rewriting, applying a rewrite $l \to r$ to an e-graphs first entails searching for instances of the left-hand pattern $l$. 
This yields matches, but now each match $\sigma$ maps pattern variables to e-classes instead of subterms. 
Then $r[\sigma]$ is added to the matched e-class, only adding information to the e-graph, instead of destructively replacing the term.
Applying the rule $x \times 2 \to x \ll 1$ in \autoref{fig:egraphs} only adds information (white e-nodes on the right).

\subsection{Equality Saturation}

A recent technique called \textit{equality saturation} \cite{eqsat, eqsat-llvm, egg}
mitigates the rewrite-choice problem by allowing rewrites to be applied simultaneously.
Na{\"i}vely generating all possible rewritings of a term would require exponential space and time, 
so equality saturation uses e-graphs to efficiently represent the massive space of equivalent terms.

Equality saturation takes in a term to optimize and a set of rewrites.
First, an initial e-graph is created from the input term.
Then, the rewrites are applied until either \textit{saturation} is reached, 
meaning the rewrites added no new information to the e-graph, or until a specified timeout.
The resulting e-graph encodes an equivalence relation over a large set of terms, all of which were built by applying rewrites to the initial term.
Finally, a procedure called \textit{extraction} selects the best represented term from the root e-class according to a user-provided cost function.
Extraction procedures can vary in their speed and the cost functions they support; 
both simple greedy algorithms \cite{herbie} and more complex ILP-powered solutions \cite{eqsat, spores} have been used.
The extracted term is guaranteed (if the rewrites themselves are sound) to be equivalent to the input term, 
and is thus returned as the optimized program.

Equality saturation effectively breaks down optimization into two phases:
the \textit{exploration} phase grows an e-graph by applying rewrites,
and then the \textit{extraction} phase selects the best term from the search space.
This decomposition avoids the rule choice problem in conventional term rewriting;
instead of having to choose which rewrite to apply to end up at the best term,
the algorithm first generates \emph{all} rewritten terms,
leaving the choice of which term to select to the extraction procedure.

Equality saturation can handle rules that could lead to non-termination in other rewriting settings, 
such as $x + y \to y + x$ and $a \to a \times 1$.
These rules may not be useful on their own, but applying them may allow other, more useful rewrites to fire.
Equality saturation allows fewer, smaller rules to compose to prove large equalities,
  where other rewriting systems have to use many larger, similar rewrite rules to avoid problematic, non-terminating rewrites.
  
Equality saturation can also prove things that directed rewriting could not, 
  even with an oracle to choose the correct rewrite at the correct time.
Consider the two rewrites $f(a, b) \to c$ and $a \to b$, and let the input term be $f(b, a)$.
Directed rewriting could only rewrite the term to $f(b, b)$, but applying $a \to b$ in a e-graph would 
prove that $f(a, b) = f(b, a) = f(a, a) = f(b, b)$, i.e., those would all be represented in the same e-class.
Once the first rewrite is applied, $c$ would also be added to the e-class.
While contrived, this example demonstrates that rewriting with e-graphs can give strictly more proving power;
in other words, if equality saturation can rewrite $e$ to $e'$ with ruleset $R$, 
there may not be a way to do the same with $R$ using traditional rewriting.

\section{\ourname{}'s Representations}

This section describes how \ourname{} represents tensor computation graphs and rewrite rules.

\subsection{Representing Tensor Computation Graphs}
\label{sec:language}

We use a representation based on the one in TASO \cite{taso}, with modifications to make it suitable for equality saturation. 
Table \ref{table:ops} shows the set of operators we consider. 
Each operator $o_i$ corresponds to a node $n_i$ in the graph; the node represents the output tensor of the operator.
The nodes corresponding to the inputs of $o_i$ are the children nodes of $n_i$. 
Each tensor computation graph is a DAG under this representation.

The formulations in equality saturation becomes simpler if a graph is single-rooted. Therefore, we combine all the final output nodes of a graph with \emph{noop}s to make the graph single-rooted. The noop nodes do not have any actual operators associated with them, and they will not be altered during the exploration phase, so there is no side effects.

\begin{table*}[t]
    \small
    {\centering
    \caption{Operators supported by \ourname. 
    There are four types for the nodes in our representation: 
    tensor type (T), string type (S), integer type (N), and tensor tuple type (TT).
    The integer type is used to represent parameters of the operators, such as stride, axis, and also padding and activation modes (by representing different modes using different integers). 
    The more complex, variable-length parameters (e.g. shape, axes permutation) are represented using the string type according to the specified formats. }
    \label{table:ops}
    \begin{tabular}{cccc}
    \toprule
        {\bf Operator} & {\bf Description} & {\bf Inputs} & {\bf Type signature} \\
    \midrule
        ewadd & Element-wise addition & input$_1$, input$_2$ & (T, T) $\rightarrow$ T  \\
        ewmul & Element-wise multiplication & input$_1$, input$_2$ & (T, T) $\rightarrow$ T \\
        matmul & Matrix multiplication & activation, input$_1$, input$_2$ & (N, T, T) $\rightarrow$ T \\
        conv $^a$ & Grouped convolution & stride$_h$, stride$_w$, pad., act., input, weight & (N, N, N, N, T, T) $\rightarrow$ T \\
        relu & Relu activation & input & T $\rightarrow$ T \\
        tanh & Tanh activation & input & T $\rightarrow$ T \\
        sigmoid & Sigmoid activation & input & T $\rightarrow$ T \\
        poolmax & Max pooling & {input, kernel$_{\{h,w\}}$, stride$_{\{h,w\}}$, pad., act.}  & (T, N, N, N, N, N, N) $\rightarrow$ T \\
        poolavg & Average pooling & {input, kernel$_{\{h,w\}}$, stride$_{\{h,w\}}$, pad., act.} & (T, N, N, N, N, N, N) $\rightarrow$ T \\
        transpose $^b$ & Transpose & input, permutation & (T, S) $\rightarrow$ T \\
        enlarge $^c$ & Pad a convolution kernel with zeros & input, ref-input & (T, T) $\rightarrow$ T \\
        concat$_n$ $^d$ & Concatenate & axis, input$_1$, \dots, input$_n$ & (N, T, \dots, T) $\rightarrow$ T \\
        split $^e$ & Split a tensor into two & axis, input & (N, T) $\rightarrow$ TT \\
        split$_0$ & Get the first output from split & input & TT $\rightarrow$ T \\
        split$_1$ & Get the second output from split & input & TT $\rightarrow$ T \\
        merge $^f$ & Update weight to merge grouped conv & weight, count & (T, N) $\rightarrow$ T \\
        reshape $^g$ & Reshape tensor & input, shape & (T, S) $\rightarrow$ T \\
        input & Input tensor & identifier $^h$ & S $\rightarrow$ T \\
        weight & Weight tensor & identifier $^h$ & S $\rightarrow$ T \\
        noop $^i$ & Combine the outputs of the graph & input$_1$, input$_2$ & (T, T) $\rightarrow$ T \\
    \bottomrule
    \end{tabular}
    }
    \captionsetup{labelformat=empty, labelsep=colon}
    \vskip 0.2em
    
    \footnotesize{$^a$ Same representation as TASO \cite{taso}. Normal and depth-wise convolutions are special cases of grouped convolutions.} \\
    \footnotesize{$^b$ Axis permutation for transpose is specified using a string with format: axis$_1$\_axis$_2$\_\dots .}\\
    \footnotesize{$^c$ Pad a convolution kernel (input) with zeros to make it the same size as ref-input. }\\
    \footnotesize{$^d$ Since each type of node needs to have a fixed number of inputs, we have a separate concat for each number of inputs. }\\
    \footnotesize{$^e$ Split the tensor in the given axis. The position of the split is at the place of the most recent concat. }\\
    \footnotesize{$^f$ Merge every \textit{count} number of groups in the grouped convolution. See TASO \cite{taso} for more details. }\\
    \footnotesize{$^g$ Specify the target shape using a string with format: dim$_1$\_dim$_2$\_\dots .}\\
    \footnotesize{$^h$ The identifier for an input or weight tensor contains its name and shape, specified as a string with format: name@dim$_1$\_dim$_2$\_\dots .}\\
    \footnotesize{$^i$ For combining the outputs of the graph to make the graph single-rooted. No actual operator is associated with noop.}
\end{table*}

\subsection{Representing Rewrite Rules}
\label{sec:rewrite}

A rewrite rule for tensor computation graph specifies that some local subgraph pattern (\textit{source pattern}) is equivalent to another subgraph pattern (\textit{target pattern}). 
The input tensors to the source and target patterns are \textit{variable nodes}, which can be substituted with any concrete nodes (or e-class in equality saturation) in the current graph. 
Each output tensor in the source pattern corresponds to an output tensor in the target pattern. 
The two corresponding output nodes are called a pair of \textit{matched outputs}. 
A rewrite rule states the equivalence between each pair of matched outputs. 

\begin{figure}[t]
    \centering
    \includegraphics[width=\linewidth]{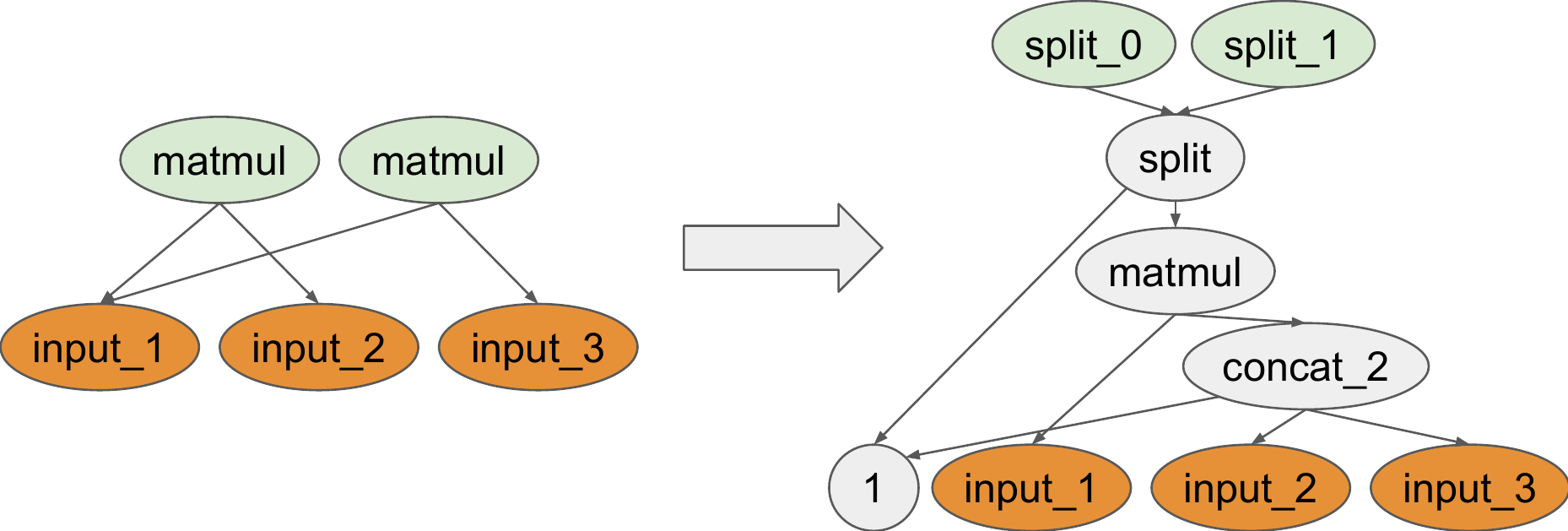}
    \begin{scriptsize}
    Source: (matmul ?input$_1$ ?input$_2$), (matmul ?input$_1$ ?input$_3$) \\
    Target: (split$_0$ (split 1 (matmul ?input$_1$ (concat$_2$ 1 ?input$_2$ ?input$_3$)))), \\
    (split$_1$ (split 1 (matmul ?input$_1$ (concat$_2$ 1 ?input$_2$ ?input$_3$))))
    \end{scriptsize}
    \caption{
    Example rewrite rule and its representation in S-expressions. 
    Identifiers starting with "?" denote variable nodes.
    For clarity, we omit the activation mode inputs to \texttt{matmul}.
    Arrows point from parent nodes to children nodes.
    1 is the axis for \texttt{split} and \texttt{concat} operators.
    }
    \label{fig:rewrite}
\end{figure}

We represent each source (and target) pattern using symbolic expressions (S-exprs) with variables. 
Patterns with a single output is represented with an S-expr rooted on the output. 
Rewrite rules with such patterns are called \textit{single-pattern rewrite rules}.
Patterns with multiple outputs are represented as a list of S-exprs rooted on each output. 
Rewrite rules with multiple matched outputs are called \textit{multi-pattern rewrite rules}.
Figure \ref{fig:rewrite} shows an example rewrite rule and its representation.

\section{\ourname{}'s Exploration Phase}
\label{sec:saturation}

We initialize the e-graph with the original tensor computation graph. 
In each iteration of the exploration phase, we search for matches of all rewrite rules in the current e-graph, and add the target patterns and equivalence relations to the e-graph. 
This process continues until either the e-graph saturates or a user-specified limit (in terms of time, e-graph size, or number of iterations) is reached.
Before applying a rewrite at a found match, we perform a \textit{shape checking} to verify if the tensor shapes in the target pattern are compatible.
This is necessary since some rewrite rules requires input tensor shapes to satisfy specific preconditions, in addition to the syntactic match. 
We perform shape checking in the same way as TASO \cite{taso}.

\paragraph{Multi-Pattern Rewrite Rules}

Multi-pattern rewrite rules are an important type of rules for tensor graph superoptimization \cite{taso}.
However, most equality saturation toolkits only support efficient search methods to find matches for single-pattern rewrite rules \cite{egg, ematching}.
We introduce an algorithm for applying multi-pattern rewrites, as shown in \cref{alg:multi}. Our algorithm leverages the existing efficient search routine for single-pattern rewrites as a subroutine.

At the beginning of the exploration phase, we collect the set of unique S-exprs present in the source patterns of the rewrite rules after canonicalization.
Here, if one S-expr can be transformed into another S-expr by variable renaming only, they will be mapped to the same canonicalized S-expr.
In each iteration of the exploration phase, we use the single-pattern search subroutine to search for matches of the canonical S-exprs. 
Then for each multi-pattern rule, we take the Cartesian product of the matches found, decanonicalize the variable-to-e-class map into the original variables (using the variable renaming map stored during canonicalization), and check if the matches are compatible at the shared variables between the S-exprs (i.e., if the shared variables refer to the same e-class after the mapping). 
We apply the matches that are compatible. 

\begin{algorithm}[t]
\small
\caption{Applying multi-pattern rewrite rules}
\label{alg:multi}
\begin{algorithmic}[1]
\Require starting e-graph $\mathcal{G}$, set of multi-pattern rewrite rules $\mathcal{R}_m$.
\Ensure updated e-graph $\mathcal{G}$.
\State canonicalized S-expr $e_c$ = Set(\{\})
\For{rule $r \in \mathcal{R}_m$ }
\For{$i = 0, \dots, |r|-1$ } \Comment{{\scriptsize $|r|$: \#S-exprs in source pattern}}
    \State ($e$, rename\_map) = \Call{Canonical}{$r$.source[$i$]}
    \State $e_c$.insert($e$)
    \State $r$.map[i] = rename\_map
\EndFor
\EndFor

  \For{iter = 0, \dots, MAX\_ITER}
    \State $M$ = \Call{Search}{$\mathcal{G}, e_c$} \Comment{all matches for all patterns}
    \For{rule $r \in \mathcal{R}_m$ }
        \For{$i = 0, \dots, |r|-1$ }
        \State canonical matches mc$_i$ = $M$[$r$.source[i]]
        \State matches m$_i$ = \Call{Decanonical}{mc$_i$, $r$.map[$i$]}
        \EndFor
        \For{$(\sigma_0, \dots, \sigma_{|r|-1}) \in$ m$_0 \times \dots \times$ m$_{|r|-1}$}
            \If{\Call{Compatible}{$(\sigma_0, \dots, \sigma_{|r|-1})$}}
            \State \Call{Apply}{$\mathcal{G}, r, \sigma_0, \dots, \sigma_{|r|-1}$}
            \EndIf
        \EndFor
    \EndFor
  \EndFor
\State \Return $\mathcal{G}$
\end{algorithmic}
\end{algorithm}

In our experience, one feature of multi-pattern rules for tensor graph is that they can grow the e-graph extremely rapidly. 
Let's consider again the example rewrite rule in Figure \ref{fig:rewrite}. 
This rule can be matched with any two \texttt{matmul} nodes with a shared input (input$_1$). 
By applying this rule once on some match, a new \texttt{matmul} node will be created and added to the e-graph (the one on the RHS of Figure \ref{fig:rewrite}), which also has input$_1$ as its input. 
If the e-graph contains $N$ \texttt{matmul} nodes that has some input$_1$ at the beginning, then after iteration 1, $\mathcal{O}(N^2)$ new \texttt{matmul} nodes sharing input$_1$ will be created. 
In iteration 2, each pair in these $\mathcal{O}(N^2)$ nodes will be a match, which will create $\mathcal{O}(N^4)$ new nodes. 
Such double exponential growth can quickly explode the e-graph.

Based on this feature, we set a separate limit $k_{\textrm{multi}}$ on the number of iterations to apply the multi-pattern rules. 
After $k_{\textrm{multi}}$ iterations, we only apply the single-pattern rules until saturation or some user-specified limit.

\section{\ourname{}'s Extraction Phase}

During extraction, the goal is to pick one e-node from each e-class in the e-graph to obtain an optimized graph. 
The optimized graph should minimize the total cost with respect to a given cost model. 
In tensor graph superoptimization, the cost model reflects the inference time taken by the graph. 

\paragraph{Cost model}
We use the same cost model as TASO \cite{taso}. 
Each operator has a separate and independent cost, which is the measured runtime of that operator (with the specific input sizes and parameters) on hardware. 
The total cost of a graph is the sum of costs of each of its nodes. 
This cost model is suitable for GPUs, since GPUs typically run one operator at a time when executing a graph \footnote{There could be more complex hardware that may execute multiple kernels in parallel, which means an accurate cost model might involve many non-local dependencies. For such cases, one may need a different extraction approach, for example using a learned approach to perform extractions. We leave this for future work.}. 
Note that an operator can be a fused operator, consisting of multiple primitive operators, such as a fused convolution and ReLU.

\subsection{Extraction Algorithms}
\label{sec:extraction}

\paragraph{Greedy extraction}

We first experiment with a greedy extraction strategy that has been shown to be effective for certain domains \cite{herbie, spores, egg}. 
For each e-class, the greedy strategy computes the total cost of the subtrees rooted on each of the e-nodes, and picks the e-node with the smallest subtree cost. 

Greedy extraction is not guaranteed to extract the graph with the minimum cost, even under our independent cost model.
For example, if two children of an e-node share a subgraph, greedy extraction would ignore the sharing and overestimate the cost. 

\paragraph{ILP extraction}
The second approach we experiment with is formulating the extraction problem as an Integer Linear Program (ILP).

Let $i = 0, ..., N-1$ be the set of e-nodes in the e-graph. 
Let $m = 0, ..., M-1$ be the set of e-classes in the e-graph.
Let $e_m$ denote the set of e-nodes within e-class $m$: $\{i | i\in e_m \}$. 
Let $h_i$ denote the set of children e-classes for e-node $i$.
Let $g(i)$ denote the e-class of e-node $i$, i.e. $i\in e_{g(i)}$.
Let $m=0$ be the root e-class. 
Each e-node is associated with a cost $c_i$. 

We then formulate our problem as follows:
\begin{align*}
    &\textrm{Minimize: } f(x) = \sum_{i} c_i x_i 
\end{align*}
Subject to:
\begin{align}
    &x_i \in \{0, 1\}, \\
    &\sum_{i\in e_0} x_i = 1, \\
    &\forall i, \forall m \in h_i, x_i \leq \sum_{j\in e_m} x_j , \\
    & \forall i, \forall m \in h_i, t_{g(i)} - t_m - \epsilon + A (1 - x_i) \geq 0, \\
    &\forall m, 0 \leq t_m \leq 1, 
\end{align}

Here we introduce a binary integer variable $x_i$ for each e-node $i$; node $i$ is selected if $x_i=1$, and not selected otherwise.
Constraint (2) ensures that one node is picked in the root e-class.
Constraint (3) ensures that if a node is picked, then at least one node in each of its children e-classes needs to be picked. 
We rely on the fact that at the optimal solution, each e-class can have at most one picked node (otherwise we can remove more picked nodes in this e-class to reduce the objective while still satisfying all the constraints). 
Constraints (1)--(3) and the objective encode the main extraction logic. 

A more subtle requirement on the extraction phase is that the extracted graph cannot contain cycles.
While the e-graph can (and likely will) contain cycles, the extracted graph is meant to map directly to an executable tensor DAG.
The extraction procedure must therefore take care to respect the acyclic invariant of DAGs.

\begin{figure}
    \centering
    \includegraphics[width=\linewidth]{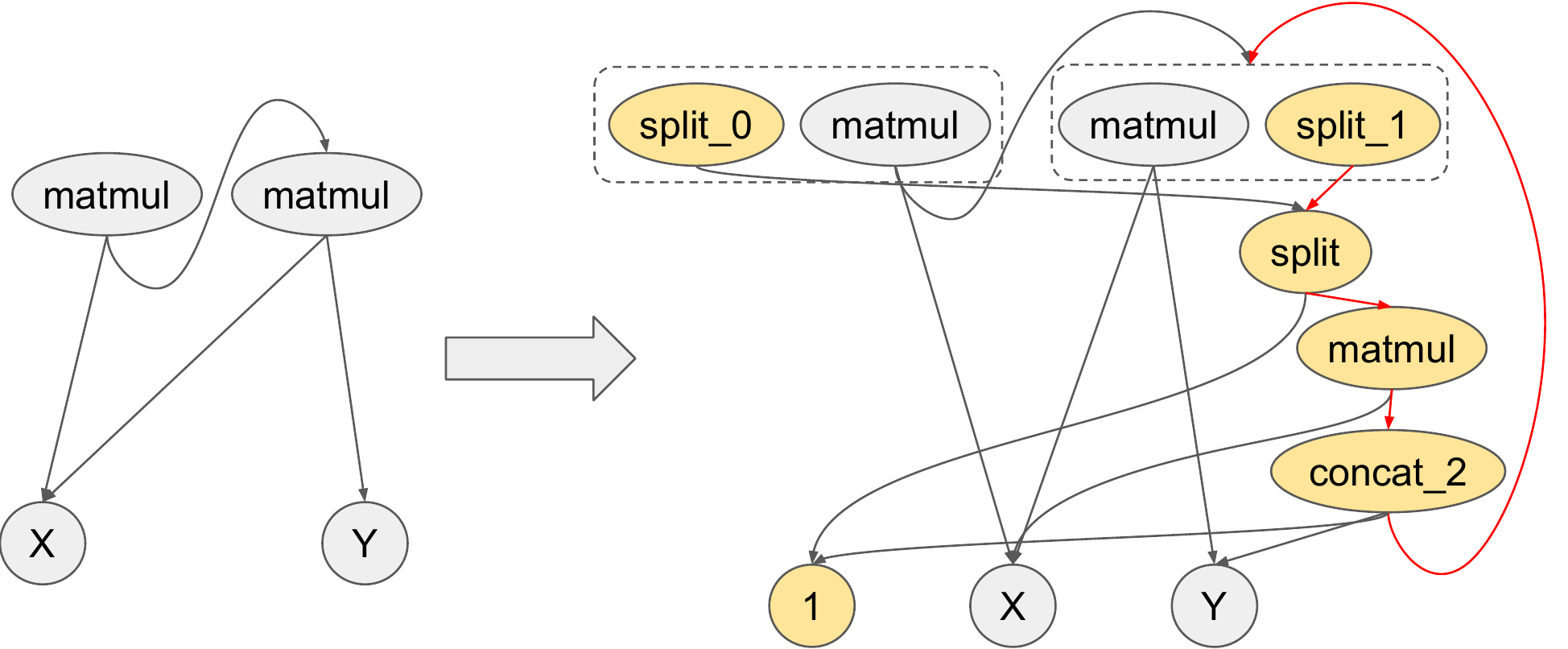}
    \vspace{-1.5em}
    \caption{Example on how a valid rewrite can introduce cycles into the e-graph. RHS is the resulting e-graph after applying the rewrite rule from Figure \ref{fig:rewrite} to the LHS. Dotted lines circles the e-classes. We omit the e-classes with a single node for clarity. If the node split$_1$ is picked in the right e-class, then the resulting graph will have a cycle (indicated by the red edges).}
    \vspace{-1em}
    \label{fig:cycle}
\end{figure}

\Cref{fig:cycle} shows an example to illustrate how valid rewrites can produce cycles in the e-graph. 
To ensure the extracted graph does not contain cycles, we introduce a real variable $t_m$ for each e-class $m$ in the ILP.
Constraint (4) ensures that the order defined by $t_m$'s is a valid topological order for the extracted graph. 
Here $\epsilon < 1/M$ is a small constant for effectively encoding strict inequalities in ILP. 
$A$ is a large enough constant such that $A > 1 + \epsilon$.
Constraint (5) is to limit the range for the topological order variables $t_m$'s.

We also experiment with using integer variables for $t_m$'s. In this case, $t_m$'s are constrained to take integer values between 0 to $M-1$. 
Constraint (4) changes accordingly to: $\forall i, \forall m \in h_i, t_{g(i)} - t_m + A (1 - x_i) \geq 1$, where $A \geq M$. 

Unlike greedy extraction, the optimal solution to the ILP is guaranteed to give a valid graph (no cycles) with the lowest cost. 

\subsection{Cycle Filtering}
\label{sec:cycle}

Similar to previous work that uses ILP extraction \cite{eqsat, spores},
we find that as the size of the e-graph grows bigger, the ILP solver takes a long time and becomes the main bottleneck.
This is mainly due to the cycle constraint (4): ILP solver struggles to find a feasible solution with these constraints. 
Therefore, we explore an alternative approach by filtering cycles during the exploration phase to make sure that the e-graph does not contain any cycles at the end of the exploration phase.
This way, we can get rid of the cycle constraints in the ILP. 

\paragraph{Vanilla cycle filtering}
The first method is to check if applying a substitution introduces cycles to the e-graph, and discard such a substitution.
This check is run every time before applying a substitution. 
Each check requires a pass over the entire e-graph. 
For one iteration during the exploration phase, this vanilla cycle filtering has complexity $\mathcal{O}(n_m N)$, where $N$ is the current size of the e-graph and $n_m$ is the total number of matches of the rewrite rules on the e-graph.

\paragraph{Efficient cycle filtering}

As the number of matches $n_m$ is typically large and scales with $N$, vanilla cycle filtering can be slow. 
We therefore design a novel and more efficient cycle filtering algorithm, consisting of a \textit{pre-filtering} step and a \textit{post-processing} step.
Algorithm \ref{alg:cycle} shows the pseudocode for the exploration phase with efficient cycle filtering. 

At the start of each iteration, we do one pass over the e-graph to record the set of descendent e-classes for each e-node (stored in a descendants map). 
During the iteration, for each match of the rewrite rules, we use the pre-stored descendants map to check if applying a rule introduces cycles to the e-graph; if so, we skip this match.
Line 3--9 implements the pre-filtering step.
Notice that this check is sound but not complete: a match that passes this check can still introduce cycles to the e-graph. 
This is because new descendants relations introduced by the previous rewrite in this iteration are not included in the pre-stored descendants map.

\begin{algorithm}[t]
\small
\caption{Exploration phase with efficient cycle filtering}
\label{alg:cycle}
\begin{algorithmic}[1]
\Require starting e-graph $\mathcal{G}$, set of rewrite rules $\mathcal{R}$.
\Ensure updated e-graph $\mathcal{G}$, filter list $l$
  \State $l = $ $\{\}$
  \For{iter = 0, \dots, MAX\_ITER}
    \State descendants map $d$ = \Call{GetDescendants}{$\mathcal{G}, l$}
    \State matches = \Call{Search}{$\mathcal{G}, \mathcal{R}, l$} 
    \For{match $\in$ matches}
      \If{\Not \Call{WillCreateCycle}{match, $d$}}
        \State \Call{Apply}{$\mathcal{G}$, match}
      \EndIf
    \EndFor
    \While{true}
      \State cycles = \Call{DfsGetCycles}{$\mathcal{G}, l$}
      \If{len(cycles) == 0}
        \State \textbf{break}
      \EndIf
      \For{cycle $\in$ cycles}
        \State \Call{ResolveCycle}{$\mathcal{G}, l$, cycle}
      \EndFor
    \EndWhile
  \EndFor
  \State \Return $\mathcal{G}, l$
\end{algorithmic}
\end{algorithm}

To resolve the cycles we missed in the pre-filtering step, we add a post-processing step at the end of each iteration (line 10-18).
We make a pass over the e-graph in DFS order and collect a set of cycles in the e-graph. 
For each cycle, we choose the last node that is added to the e-graph, and add that node to a filter list. 
The nodes in the filter list are considered as removed from the e-graph. 
We make sure those nodes are not picked during extraction by explicitly adding constraints $\forall i \in l, x_i = 0$ to the ILP.

By constructing a descendants map once before each iteration, each of the checking in the pre-filtering step takes constant time. 
The worst case complexity of the post-processing step is $\mathcal{O}(n_c N)$, where $n_c$ is the number of cycles in the e-graph.
Since $n_c$ is typically much smaller than $n_m$, this algorithm is much faster than the vanilla cycle filtering. 
In practice, each DFS pass over the e-graph can find many cycles, making $\mathcal{O}(n_c N)$ a very conservative upper bound.

\section{Evaluation}

We implement \ourname{} in \citet{rust} using \texttt{egg} \cite{egg}, an open source equality saturation library.
For the extraction phase, we use SCIP \cite{scip} as the ILP solver, wrapped by Google OR-tools \cite{ortools}. 

We utilize egg's \textit{e-class analysis} feature for the shape checking discussed in Section \ref{sec:saturation}. 
An e-class analysis associates data with each e-class to support rewrites that are not purely syntactic. 
We store all the relevant information of the tensors (shape, layout, split locations) in the analysis data and use these information for shape checking. 

\subsection{Experimental Setup}

We compare \ourname{} with TASO \cite{taso} to evaluate our equality saturation based search.
We use the same set of rewrite rules as TASO for our experiments \footnote{ \ourname{} supports flexible choices of rewrite rules. All rewrite rules from TASO are supported by \ourname{}. \texttt{egg} (the equality saturation library we use) also supports conditional rewrite rules (applying a rule only when a condition is met).}. 
We evaluate on the inference graphs of 7 models:
\textbf{BERT} \cite{bert}, \textbf{ResNeXt-50} \cite{resnext50}, \textbf{NasNet-A} \cite{nasneta}, \textbf{NasRNN} \cite{nasrnn}, \textbf{Inception-v3} \cite{inceptionv3},
\textbf{VGG-19} \cite{vgg}, and \textbf{SqueezeNet} \cite{squeezenet}. 
This benchmark set covers a wide range of commonly used state-of-the-art models, including both models for computer vision tasks and models for NLP tasks, both human-designed models and automatically-discovered models by neural architecture search.
We perform all experiments on a Google Cloud instance with one NVIDIA Tesla T4 GPU, a 16-core CPU, and 60 GB of memory.
We also experiment with ResNet-50 \cite{resnet}, but find that on T4 GPU, the rewrite rules from TASO cannot provide any speedup to the graph.

For \ourname{}, our full approach uses the efficient cycle filtering algorithm (Section \ref{sec:cycle}) during the exploration phase and the ILP method without the cycle constraints (Section \ref{sec:extraction}) for extraction. 
We set a limit on the number of nodes in the e-graph $N_{\textrm{max}}=50000$ and the number of iterations for exploration $k_{\textrm{max}}=15$.
We terminate the exploration phase when any of the limit is reached, or the e-graph is saturated. 
We set a separate limit $k_{\textrm{multi}}$ on the number of iterations to apply the multi-pattern rules.
We use a default of $k_{\textrm{multi}}=1$ for the main results in Section \ref{sec:speedup} and \ref{sec:time}, and study the effect of varying $k_{\textrm{multi}}$ in Section \ref{sec:multi-vary}.
We set a timeout of 1 hour for the ILP solver.

\begin{figure*}
\begin{minipage}[t]{0.36\textwidth}
    \centering
    \vspace{0pt}
    \includegraphics[width=\linewidth]{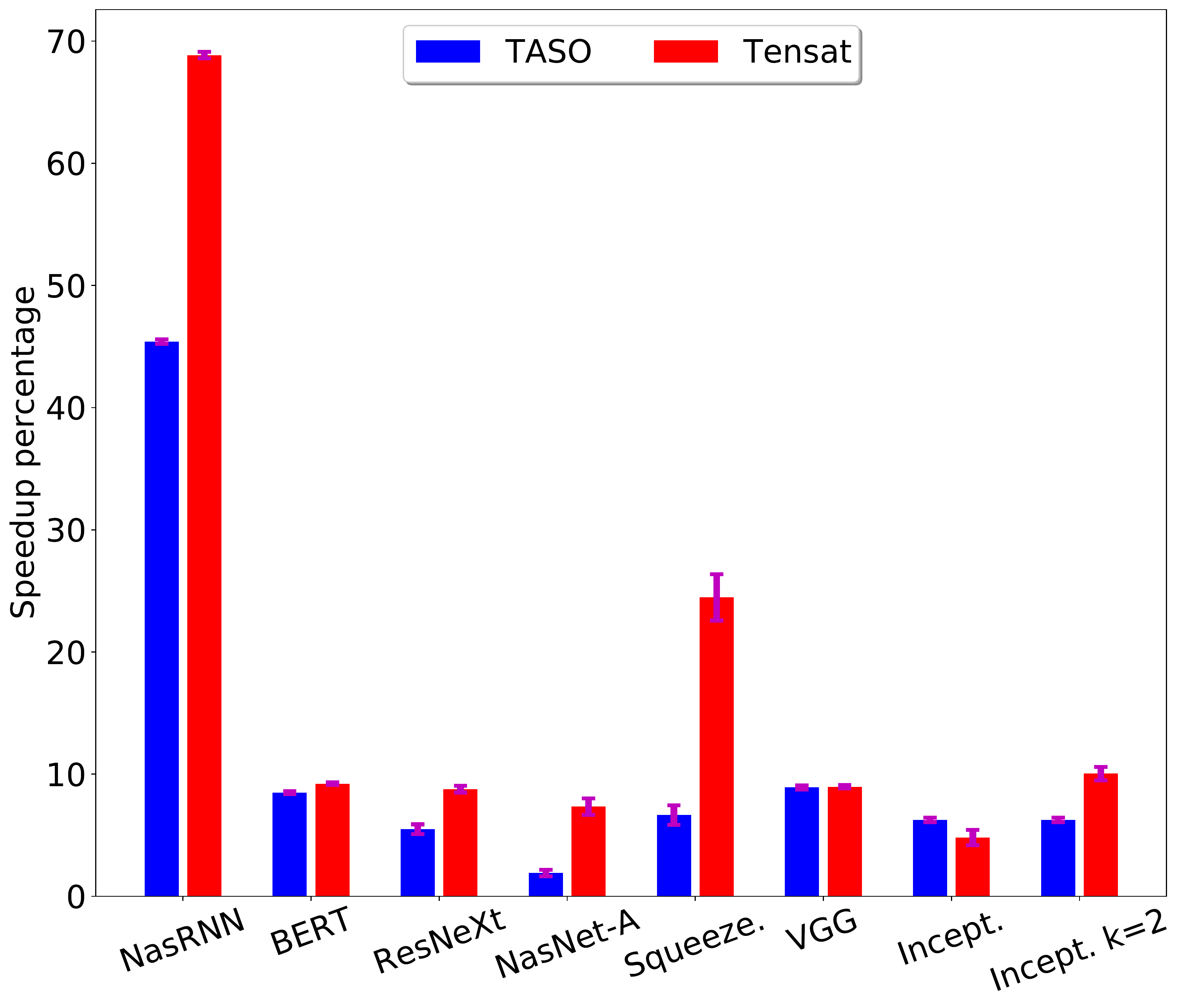}
    \caption{
        Speedup percentage of the optimized graph with respect to the original graph: TASO v.s. \ourname{}. 
        Each setting (optimizer $\times$ benchmark) is run for five times, and we plot the mean and standard error for the measurements. 
    }
    \label{fig:speedup}
\end{minipage}
\hspace{0.01\linewidth}
\begin{minipage}[t]{0.36\textwidth}
    \centering
    \vspace{0pt}
    \includegraphics[width=\linewidth]{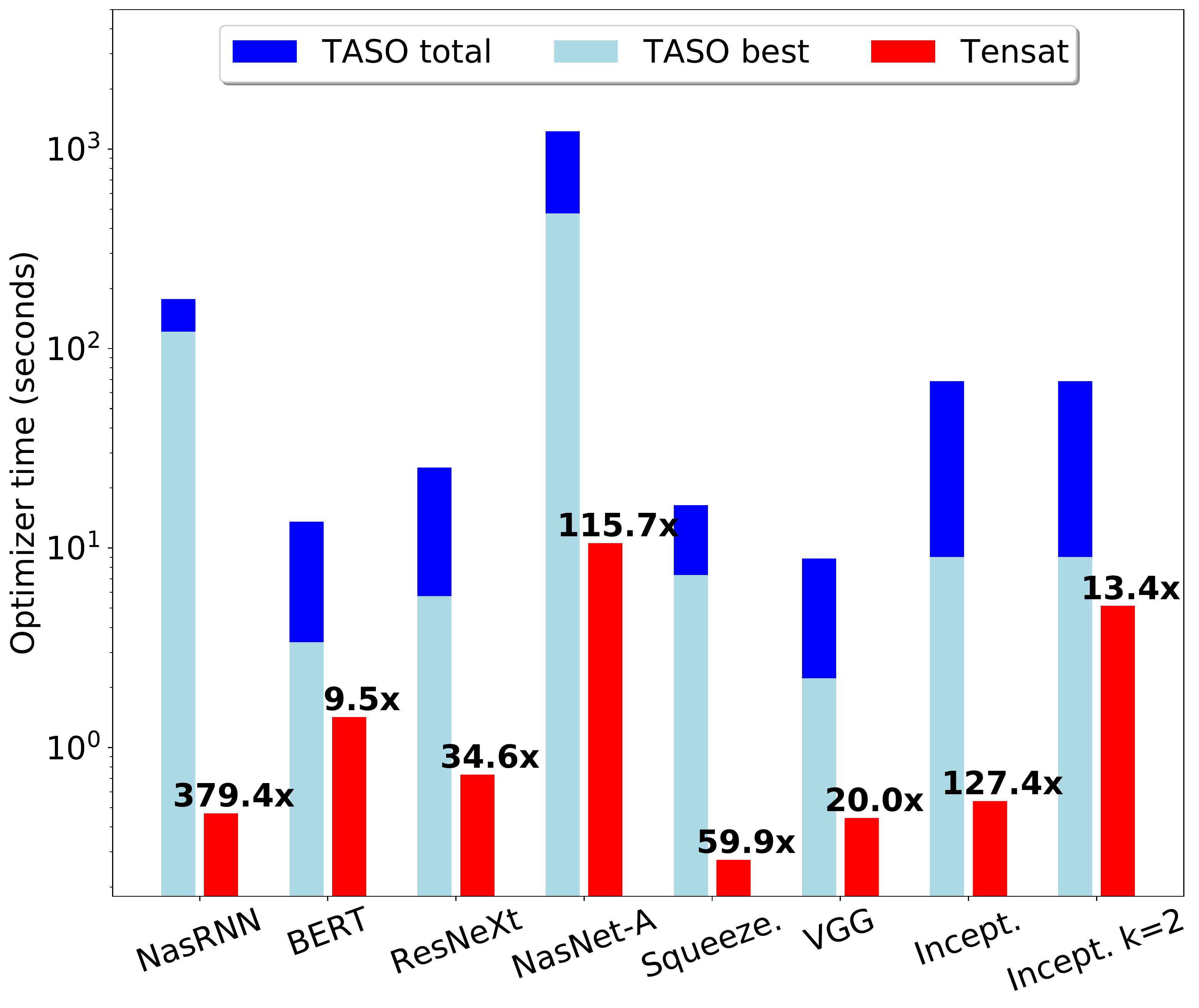}
    \vspace{-1.0em}
    \caption{
        Optimization time (log scale): TASO v.s. \ourname{}. 
        ``TASO total'' is the total time of TASO search.
        ``TASO best'' indicates when TASO found its best result;
        achieving this time would require an oracle telling it when to stop.
    } 
    \label{fig:overhead}
\end{minipage}
\hspace{0.01\linewidth}
\begin{minipage}[t]{0.23\textwidth}
\centering
\vspace{0pt}
\includegraphics[width=\linewidth]{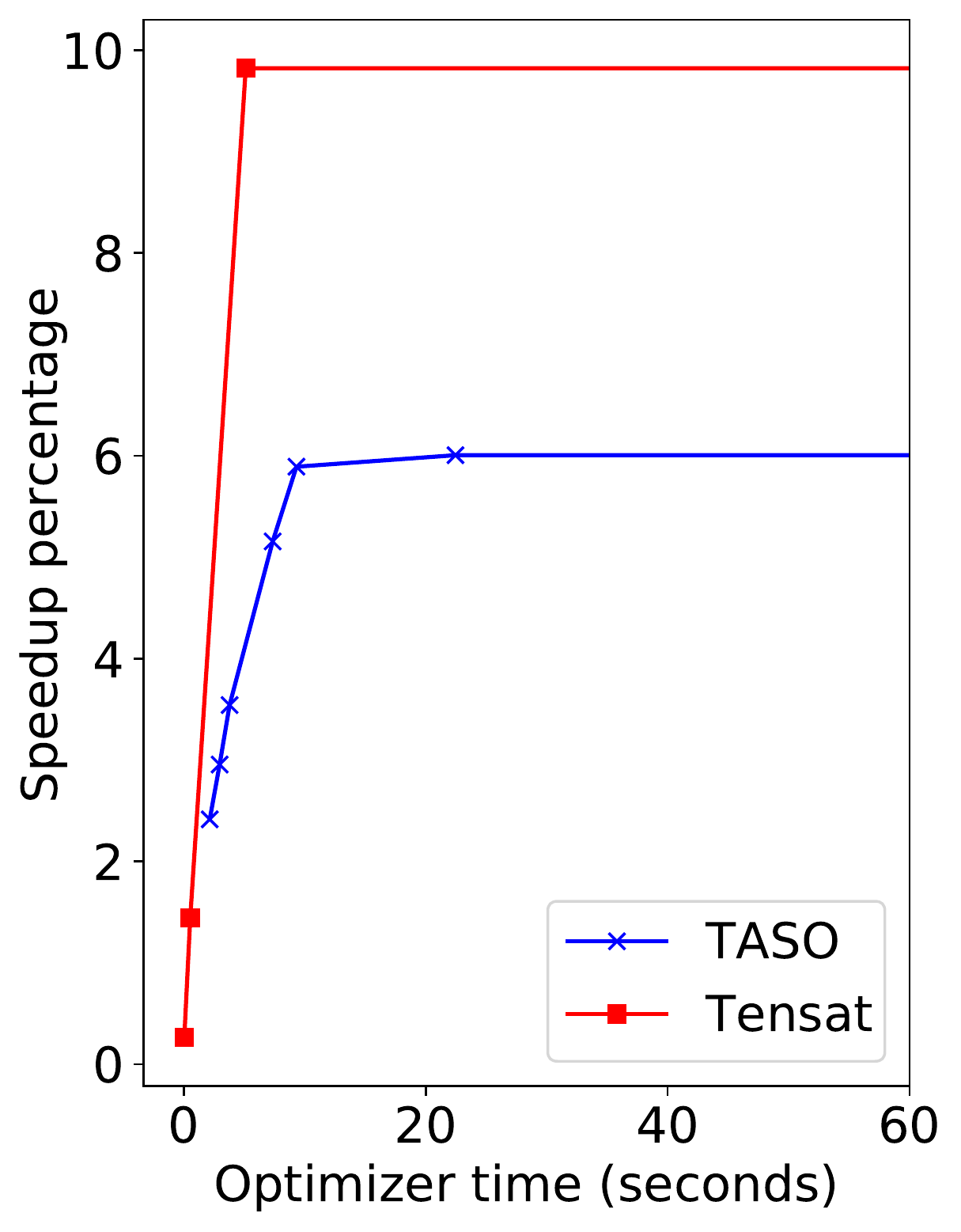}
\vspace{-0.6em}
\caption{Speedup over optimization time for TASO and \ourname{}, on Inception-v3. We use a timeout of 60 seconds.
}
\label{fig:traj}
\end{minipage}
\end{figure*}

For TASO's backtracking search, we use their default settings from their artifact evaluation code on the number of iterations\footnote{The number of iterations of the outer loop, see Algorithm 2 in \cite{taso} for more details.} $n=100$ and the hyperparameter $\alpha=1.0$ for each benchmark. 
We also test $\alpha=1.05$ as mentioned in their paper, and find that the difference is tiny (difference in speedup percentage is less than 0.1\% on average over the benchmarks).
Increasing to $n=1000$ leads to less than 1\% speedup gain with the cost of over 11x longer in optimization time on average.

\subsection{Program Speedup}
\label{sec:speedup}

We compare the speedup percentage of the optimized graph with respect to the original graph between \ourname{} and TASO. 
We use TASO's cuDNN backend to measure the runtime of the full computation graphs. 
Figure \ref{fig:speedup} shows the results. 
We can see that \ourname{} discovers better optimized graphs compared with TASO's backtracking search in most benchmarks, with the largest speedup improvements on NasRNN and SqueezeNet.
This improvement comes from the fact that equality saturation covers a much larger space of equivalent graphs than sequential backtracking search.
By using e-graph as a compact representation of an exponential number of equivalent graphs, \ourname{} is able to cover orders of magnitude more equivalent graphs than TASO.

Note that for Inception-v3, \ourname{} with $k_{\textrm{multi}}=1$ gives a smaller speedup than TASO, 
but increasing $k_{\textrm{multi}}$ to 2 achieves a better speedup than TASO
while still being 13.4$\times$ faster than TASO's search (see Figure \ref{fig:overhead}).
In general, an optimizer can achieve a better speedup given longer optimization time. 
Therefore, to compare the performance between TASO and \ourname{} on Inception-v3 more clearly, we plot the tradeoff curves of how speedup varies with optimization time in Figure \ref{fig:traj}.
We can see that \ourname{} achieves a better tradeoff curve.

We inspect the optimized graphs from \ourname{} and recorded some rewrite patterns that is used in them.
We present several examples of useful patterns in the Appendix. 

\subsection{Optimization Time}
\label{sec:time}

Another important metric is the time taken by the optimizer itself. 
For \ourname{}, this is the sum of time of the exploration phase and the extraction phase. 
For TASO, we record two times:
the first is the total time of the backtracking search with the default number of iterations ($T_{\textrm{total}}$); 
the second is the time taken to first reach the best graph found during its search ($T_{\textrm{best}}$). 
$T_{\textrm{best}}$ is the best possible time for TASO's sequential search. 
In practice, it is difficult (if not impossible) to achieve $T_{\textrm{best}}$ since the sequential search algorithm would have no way to know that it can stop at that point. 

Figure \ref{fig:overhead} shows the time taken by the optimizers across benchmarks. 
We can see that \ourname{} runs 9.5x to 379x faster than TASO's $T_{\textrm{total}}$, and 1.8x to 260x times faster than $T_{\textrm{best}}$. 
This shows that \ourname{} can not only cover a much larger search space, but also achieve this in drastically less time. 
Furthermore, \ourname{}'s optimization time is small enough that we believe our approach can be integrated into a default compilation flow instead of running the search as an additional offline autotuning process.
Table \ref{table:breakdown} shows the optimization time breakdown for \ourname{}.

\begin{table}
\centering
    \footnotesize
    \begin{tabular}{ccc}
    \toprule
   {\bf Time (s)} & {\bf Exploration} & {\bf Extraction} \\
    \midrule
    NasRNN & 0.10 & 0.37 \\
    BERT & 0.83 & 0.60 \\
    ResNeXt-50 & 0.68 & 0.05 \\
    NasNet-A & 8.81 & 1.79 \\
    SqueezeNet & 0.16 & 0.12 \\
    VGG-19 & 0.42 & 0.02 \\
    Inception-v3 & 4.38 & 0.75 \\
    \bottomrule
    \end{tabular}
    \captionof{table}{Optimization time breakdown for \ourname{}.}
    \label{table:breakdown}
\end{table}

\subsection{Varying Iterations of Multi-Pattern Rewrites}
\label{sec:multi-vary}

\begin{figure*}
    \centering
    \includegraphics[width=0.29\hsize]{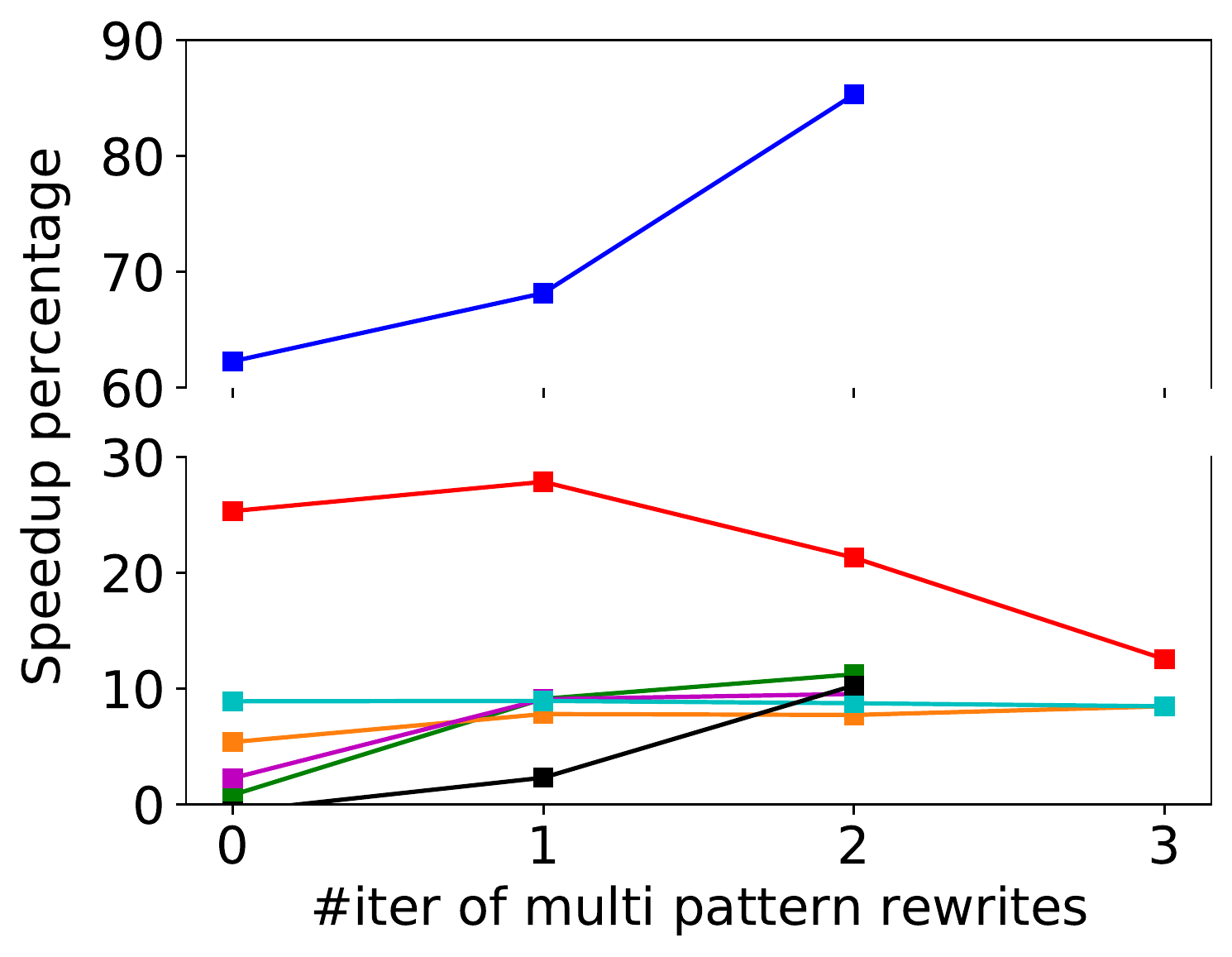}
    \includegraphics[width=0.29\hsize]{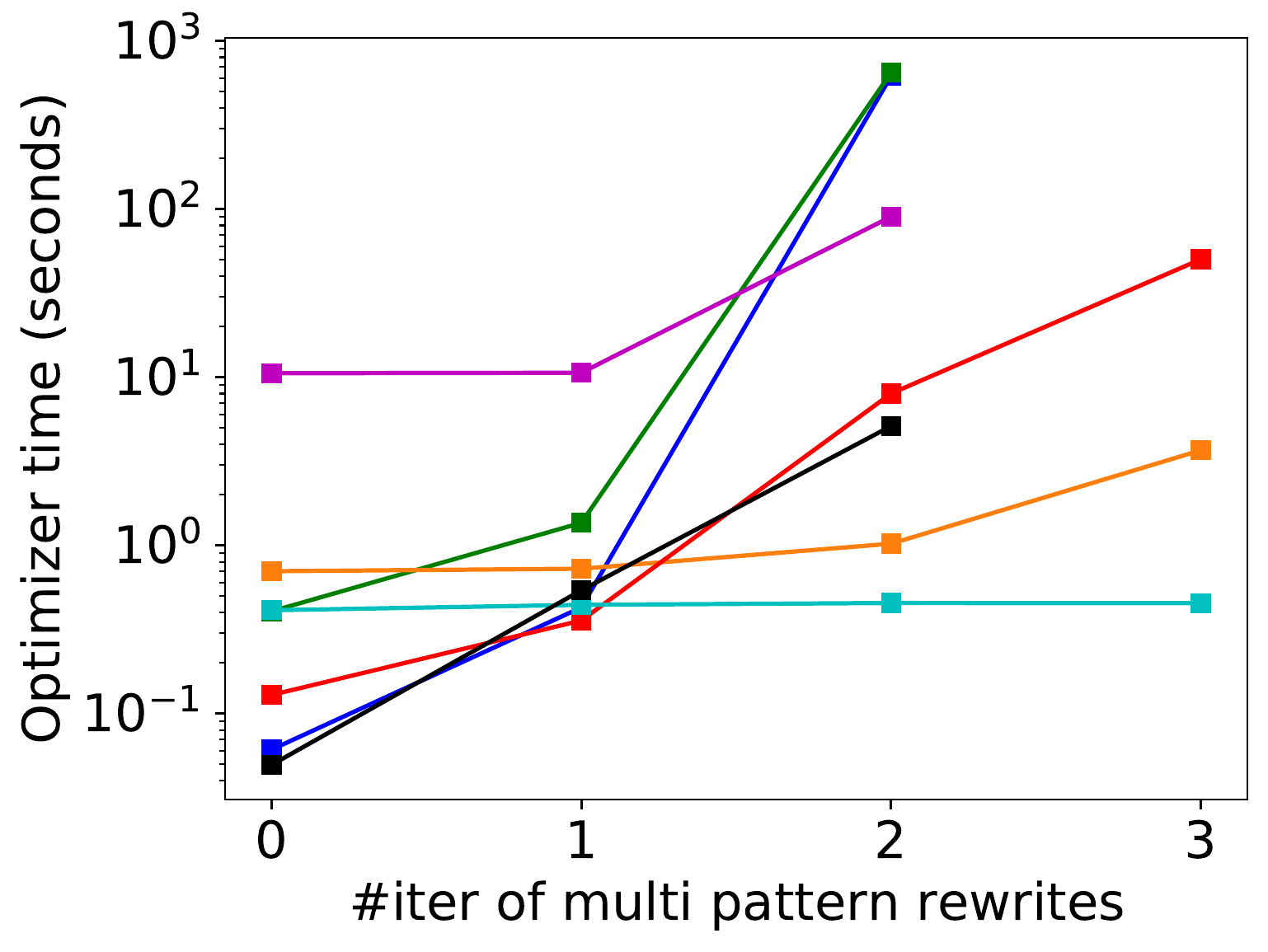}
    \includegraphics[width=0.29\hsize]{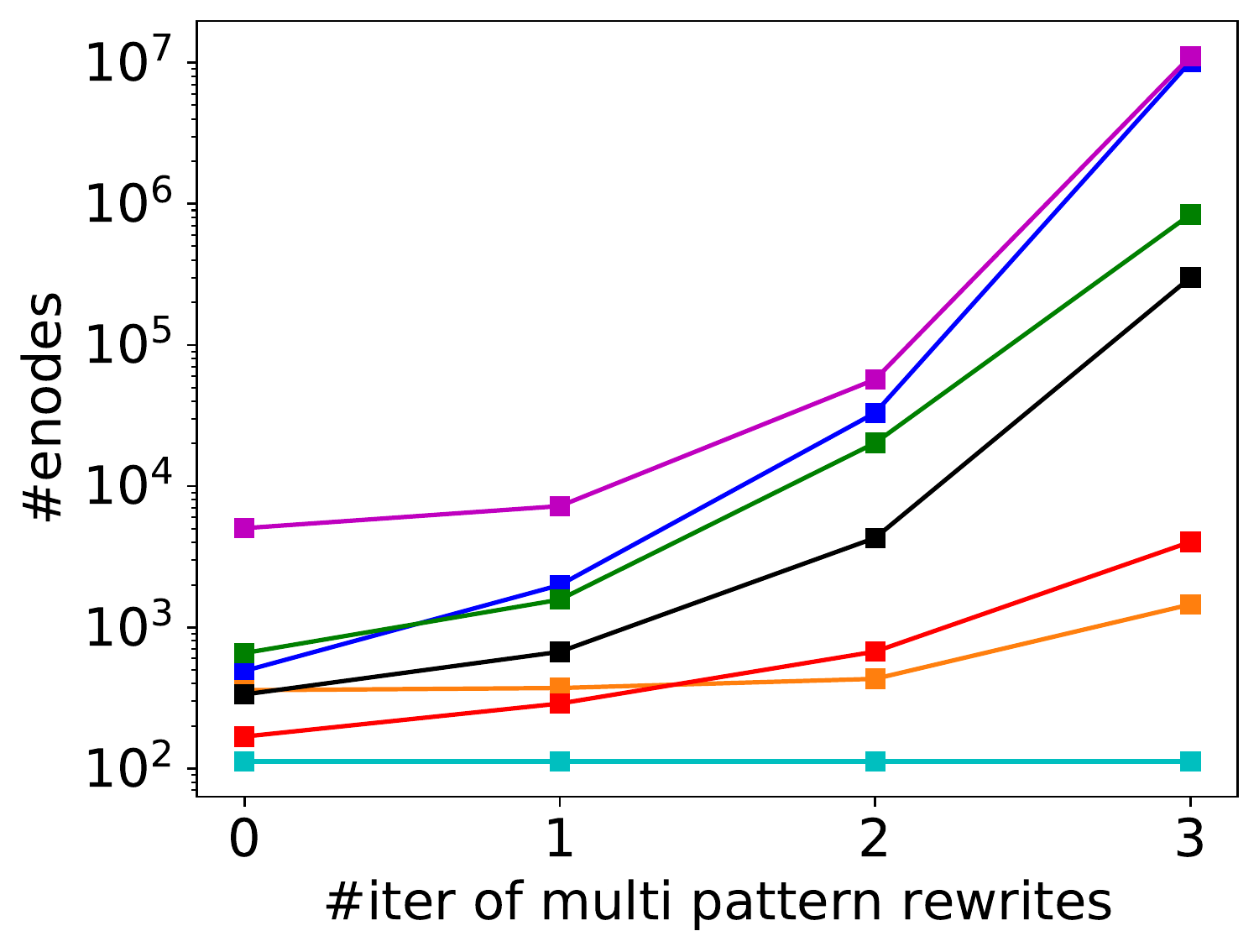}
    \includegraphics[width=0.09\hsize]{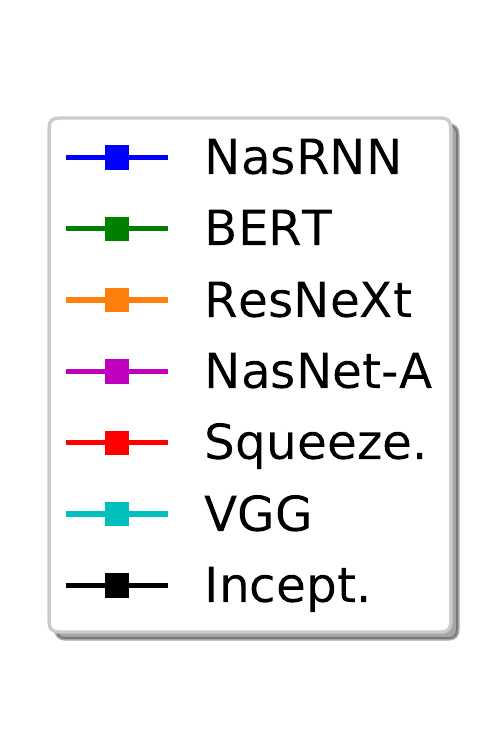}
    \vspace{-.5em}
    \caption{Effect of varying the number of iterations of multi-pattern rewrites $k_{\textrm{multi}}$. 
    For BERT, NasNet-A, NasRNN, Inception-v3, the ILP solver times out at one hour for $k_{\textrm{multi}}=3$. 
    Left: speedup of the optimized graphs (the $y$-axis is split for clarity). 
    Middle: time taken by \ourname{}. 
    Right: final e-graph size (number of e-nodes).
    The middle and right figures are in log scale.
    }
    \label{fig:trend}
\end{figure*}

As we discuss in Section \ref{sec:saturation}, multi-pattern rewrite rules can grow the e-graph in a extremely rapid manner. 
Here we study the effect of varying the number of iterations for multi-pattern rewrites $k_{\textrm{multi}}$.
Figure \ref{fig:trend} shows the results. 
We can see the explosion of the number of nodes in the e-graph as $k_{\textrm{multi}}$ increases (due to the double exponential growth). 
For NasRNN, Inception-v3, BERT, NasNet-A, and ResNeXt-50, by increasing $k_{\textrm{multi}}$, \ourname{} discovers better graphs with larger speedups. 
But for SqueezeNet, speedup decreases with $k_{\textrm{multi}}$.
This is due to the discrepancy between the cost model and the real graph runtime.
As $k_{\textrm{multi}}$ increases for SqueezeNet, the cost model suggests that certain new rewrites can reduce the cost, while they in fact increase full graph runtime. 
Despite this special case where the discrepancy has an effect, \ourname{} on SqueezeNet with $k_{\textrm{multi}}=3$ still achieves a better speedup than TASO.
By increasing $k_{\textrm{multi}}$, \ourname{} can explore a larger search space and find better optimized graphs for most benchmarks, at the cost of longer time taken by the optimizer. 

For larger and more complex networks, the current technique can only explore up to a certain number of iterations of multi-pattern rewrites, beyond which the e-graph becomes too big for the extraction phase. 
A potential approach to further improve the scalability of \ourname{} is to selectively apply rules during exploration that are more likely to provide speedup. 
One potential direction for future work is to explore using ML techniques to learn the decisions on when and where to apply certain rules.

\subsection{Ablation Study}
\label{sec:ablation}

In this section, we study the effect of the important design choices in our approach.

\paragraph{Greedy v.s. ILP extraction}
Table \ref{table:extraction} shows the comparison between greedy extraction and ILP extraction. 
Although greedy extraction works fine on some benchmarks (e.g. NasRNN), it fails to extract an optimized graph on others (e.g. BERT and NasNet-A). 
This is due to the nature of greedy extraction: it makes the choices on which node to pick separately and greedily, without considering the inter-dependencies between the choices.
Consider the rewrite in Figure \ref{fig:rewrite} (merging two \texttt{matmul}s by \texttt{concat} and \texttt{split}) as an illustrative example. 
After applying this rewrite to the e-graph, there will be two e-classes that have multiple e-nodes: one e-class per each output. 
This rewrite can reduce the cost only if both e-classes choose the split node, since the RHS subgraph can be reused by the two outputs. 
However, greedy extraction will never pick the split nodes, since it does not know the RHS subgraph is shared between the two split nodes.

\begin{table}[]
    \centering
    \footnotesize
    \begin{tabular}{cccc}
    \toprule
        {\bf Graph Runtime (ms)} & {\bf Original} & {\bf Greedy} & {\bf ILP} \\
    \midrule
        BERT & 1.88 & 1.88 & \textbf{1.73} \\
        NasRNN & 1.85 & 1.15 & \textbf{1.10} \\
        NasNet-A & 17.8 & 22.5 & \textbf{16.6} \\
    \bottomrule
    \end{tabular}
    \caption{Comparison between greedy extraction and ILP extraction, on BERT, NasRNN, and NasNet-A. 
    This table shows the runtime of the original graphs and the optimized graphs by greedy extraction and ILP extraction. 
    The exploration phase is run with $k_{\textrm{multi}} = 1$. }
    \label{table:extraction}
\end{table}

\paragraph{ILP with or without cycle constraints}

Here we study the effect of whether or not to include the cycle constraints in ILP. 
Table \ref{table:cycle} presents the effect on extraction time as $k_{\textrm{multi}}$ (thus e-graph size) varies. 
With the cycle constraints, ILP solver time quickly increases with the e-graph size, and reaches timeout when $k_{\textrm{multi}}=2$. 
In our experiments, the ILP solver has not yet found a feasible solution at timeout. 
Removing the cycle constraints leads to approximately 10x--1000x speedup on ILP solving time on larger e-graphs.
These results show that the main difficulty for the ILP solver is to satisfy the cycle constraints.  
Thus, removing the cycle constraints makes it possible for our approach to scale to larger e-graphs. 

\begin{table}[]
    \centering
    \footnotesize
    \begin{tabular}{ccccc}
    \toprule
        {\bf Extraction} & \multirow{2}{*}{\bf $k_{\textrm{multi}}$} & \multicolumn{2}{c}{\bf With cycle} & {\bf Without} \\
        {\bf time (s)} & & real & int & {\bf cycle} \\
    \midrule
       \multirow{2}{*}{BERT} & 1 & 0.96 & 0.98 & \textbf{0.16} \\
       & 2 & $>$3600 & $>$3600 & \textbf{510.3} \\
       \midrule
       \multirow{2}{*}{NasRNN} & 1 & 1116 & 1137 & \textbf{0.32}  \\
       & 2 & $>$3600 & $>$3600 & \textbf{356.7}  \\
       \midrule
       \multirow{2}{*}{NasNet-A} & 1 & 424 & 438 & \textbf{1.81}  \\
       & 2 & $>$3600 & $>$3600 & \textbf{75.1}  \\
    \bottomrule
    \end{tabular}
    \caption{Effect of whether or not to include cycle constraints in ILP on extraction time (in seconds), on BERT, NasRNN, and NasNet-A. 
    For the cycle constraints, we compare both using real variables and using integer variables for the topological order variables $t_m$. }
    \label{table:cycle}
\end{table}

\paragraph{Efficient cycle filtering}

To remove the cycle constraints from ILP, we need to perform cycle filtering during the exploration phase. 
Here we compare the two cycle filtering techniques introduced in Section \ref{sec:cycle}. 
Table \ref{table:efficient} shows the effect on the exploration phase time, as $k_{\textrm{multi}}$ varies.
We can see that the efficient cycle filtering algorithm achieves up to 2000x speedup compared with the vanilla algorithm, making it possible to explore a larger e-graph. 

\begin{table}[]
    \centering
    \footnotesize
    \begin{tabular}{cccc}
    \toprule
        {\bf Exploration time (s)} & {\bf $k_{\textrm{multi}}$} & 
        {\bf Vanilla} & {\bf Efficient} \\
    \midrule
       \multirow{2}{*}{BERT} & 1 & 0.18 & \textbf{0.17} \\
       & 2 & 32.9 & \textbf{0.89} \\
       \midrule
       \multirow{2}{*}{NasRNN} & 1 & 1.30 & \textbf{0.08}  \\
       & 2 & 2932 & \textbf{1.47} \\
       \midrule
       \multirow{2}{*}{NasNet-A} & 1 & 3.76 & \textbf{1.27} \\
       & 2 & $>$3600 & \textbf{8.62} \\
    \bottomrule
    \end{tabular}
    \caption{Comparison between vanilla cycle filtering and efficient cycle filtering, on the exploration phase time (in seconds) for BERT, NasRNN, and NasNet-A.}
    \label{table:efficient}
\end{table}

\section{Related Work}

\paragraph{Graph Rewrite Optimizations}

Our work improves the search mechanism for finding the most optimal tensor graph substitutions upon existing work \cite{taso,metaflow,Fang:sampling}. TASO \cite{metaflow,taso} uses a backtracking search algorithm with a hard threshold for allowing substitutions that increase runtime. Compared to TASO, a subsequent work \cite{Fang:sampling} presents a more efficient sampling-based search algorithm that prunes redundant substitutions. While the sampling-based approach is faster than TASO, it does not lead to discovering more optimized programs, unlike our approach. 

An optimization via graph substitutions is also critical to other domains outside deep learning. NeuRewriter \cite{NeuRewriter} exploits reinforcement learning to iteratively select which rewrite rule to apply on which region of the graph for multiple problem domains, including algebraic expression simplification, job scheduling, and vehicle routing. This technique is complement to our approach. In particular, we can enhance our approach by applying machine learning to select more promising multi-pattern rules to apply in each iteration when we cannot apply multi-pattern rules to saturation.

Unlike our approach, these prior techniques suffer from the reliance on iteratively applying substitutions in sequences.

\paragraph{Superoptimization} 
Superoptimization is a program optimization technique that searches for a correct and optimal program with respect to a cost model. Most superoptimizers optimize relatively short sequences of low-level instructions \cite{massalin,denali,Bansal:peephole,stoke:asplos,stoke:pldi,greenthumb-asplos,greenthumb-cc,Souper}. While most of them do not rely on rewrite rules, Denali \cite{denali} takes the approach of using rewrite rules for scalability but sacrificing some optimality guarantee. Similar to ours, Denali \cite{denali} employs e-graphs and extracts optimal programs using a constrain solver. 
Unlike our work, none of these prior research focuses on tensor graph superoptimization.

\paragraph{Equality Saturation Applications}
The core of \ourname{}'s approach is the application of equality saturation, which has been successfully applied in other domains as well.
The first works \cite{eqsat, eqsat-llvm} focused on traditional compiler optimizations, but more recent applications include CAD simplification, numerical accuracy, and code search \cite{szalinski, herbie, yogo}.

\citet{spores} also use equality saturation to optimize machine learning programs. 
However, they optimize linear algebra kernels consisting of few simple
operations like matrix multiplication and summation, whereas \ourname{} optimizes at the 
computation graph level.
We contribute the multi-pattern extension to
better explore the search space of deep learning models, as well as the cycle-filtering
algorithm to make ILP extraction efficient. 


\section{Conclusions}
We have presented a new approach to tensor graph optimization using equality saturation. 
We explained necessary extensions to equality saturation to make it work for our problem domain: supporting multi-pattern rewrite rules, and introducing a new extraction algorithm with an efficient cycle filtering for scalability.
We show that our approach can find optimized graphs with up to 16\% speedup over state-of-the-art, while spending on average 48x less time optimizing.
Our approach is able to find graphs that are globally optimal, and is fast enough that it can be integrated into the normal compilation flow for inference graphs.

\section*{Acknowledgments}

We thank Zhihao Jia, Martin Maas, Hyeontaek Lim, and the anonymous reviewers for their insightful and helpful comments.

\bibliography{main}
\bibliographystyle{mlsys2021}

\newpage
\clearpage
\appendix

\section{Example Patterns of Rewrite Used by \ourname{}}

\begin{figure}[h]
    \centering
    \hfill
    \includegraphics[width=.49\linewidth]{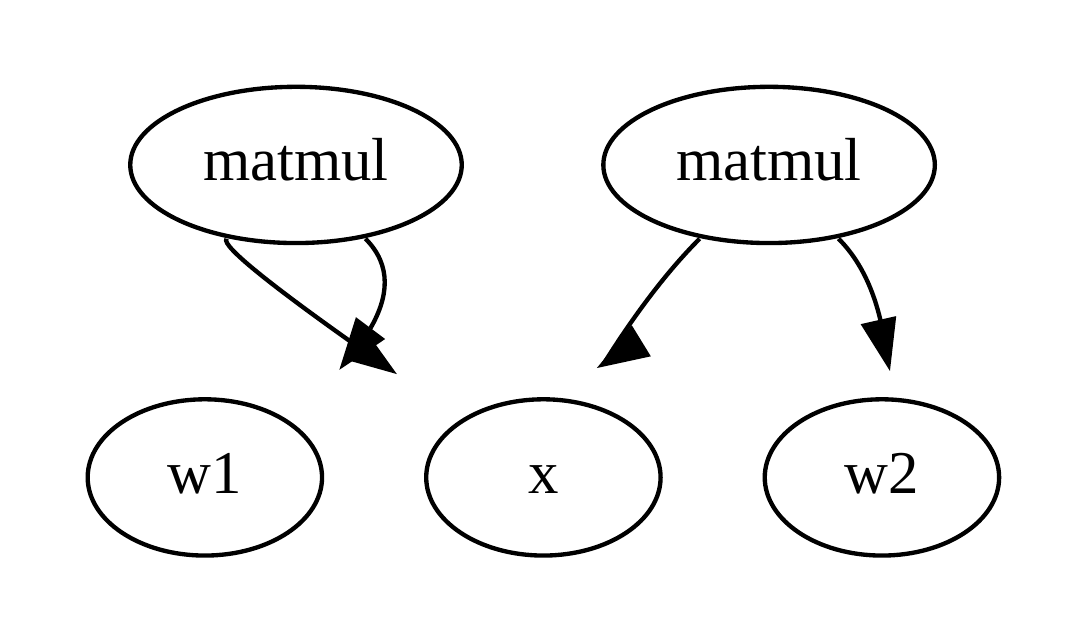}
    \hfill
    \includegraphics[width=.49\linewidth]{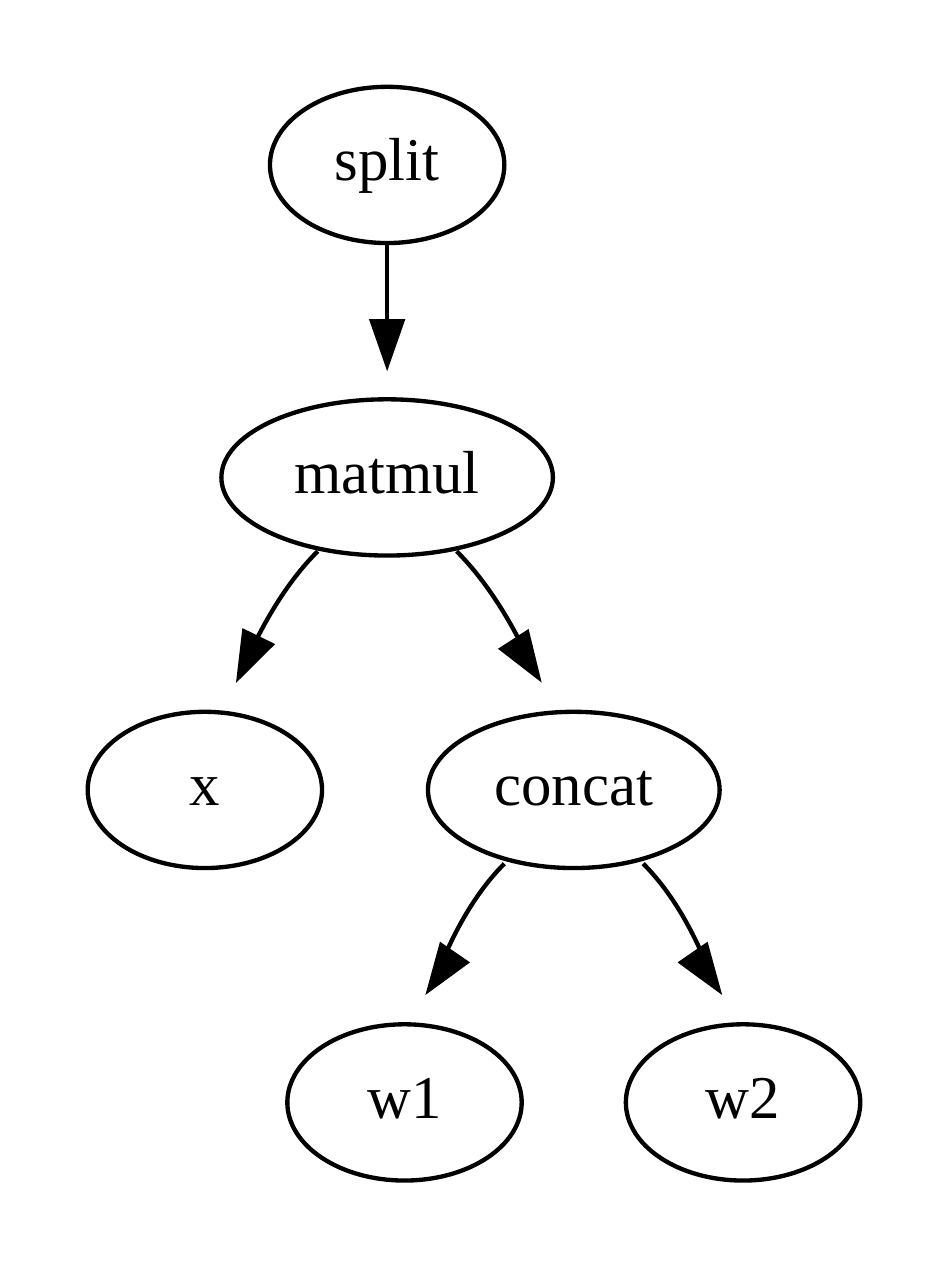}
    \hfill
    \caption{Example pattern used in BERT. 
    \texttt{w1} and \texttt{w2} are weight nodes. 
    The optimized graphs for BERT also contain this pattern generalizing to more than 2 \texttt{matmul}s sharing a common input node.}
    \label{fig:bert-rewrite}
\end{figure}

\begin{figure}[h]
    \centering
    \hfill
    \includegraphics[width=0.49\linewidth]{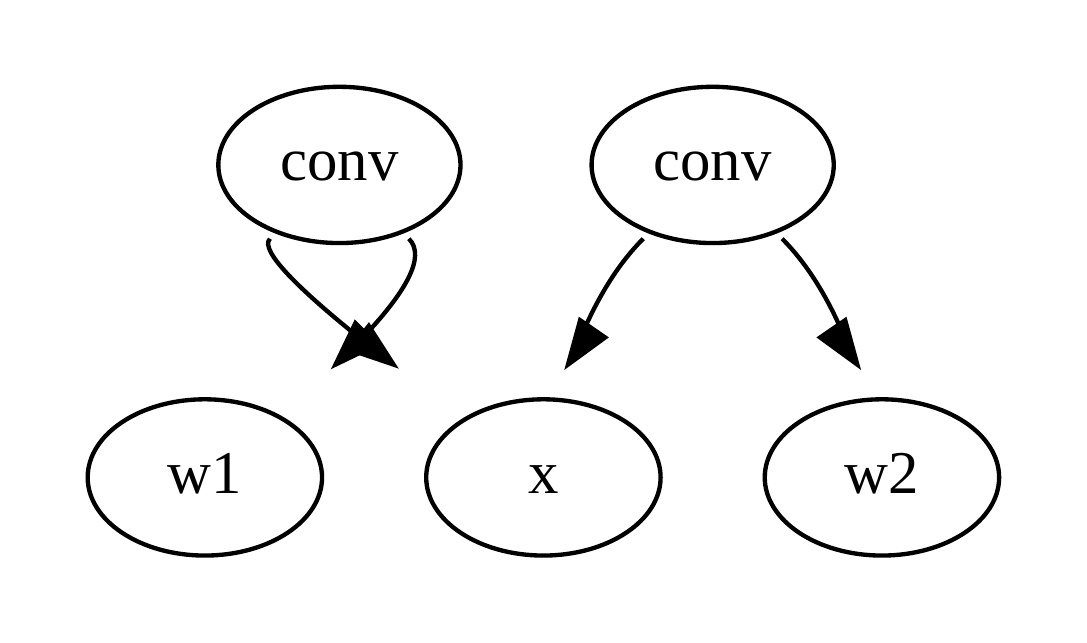}
    \hfill
    \includegraphics[width=0.49\linewidth]{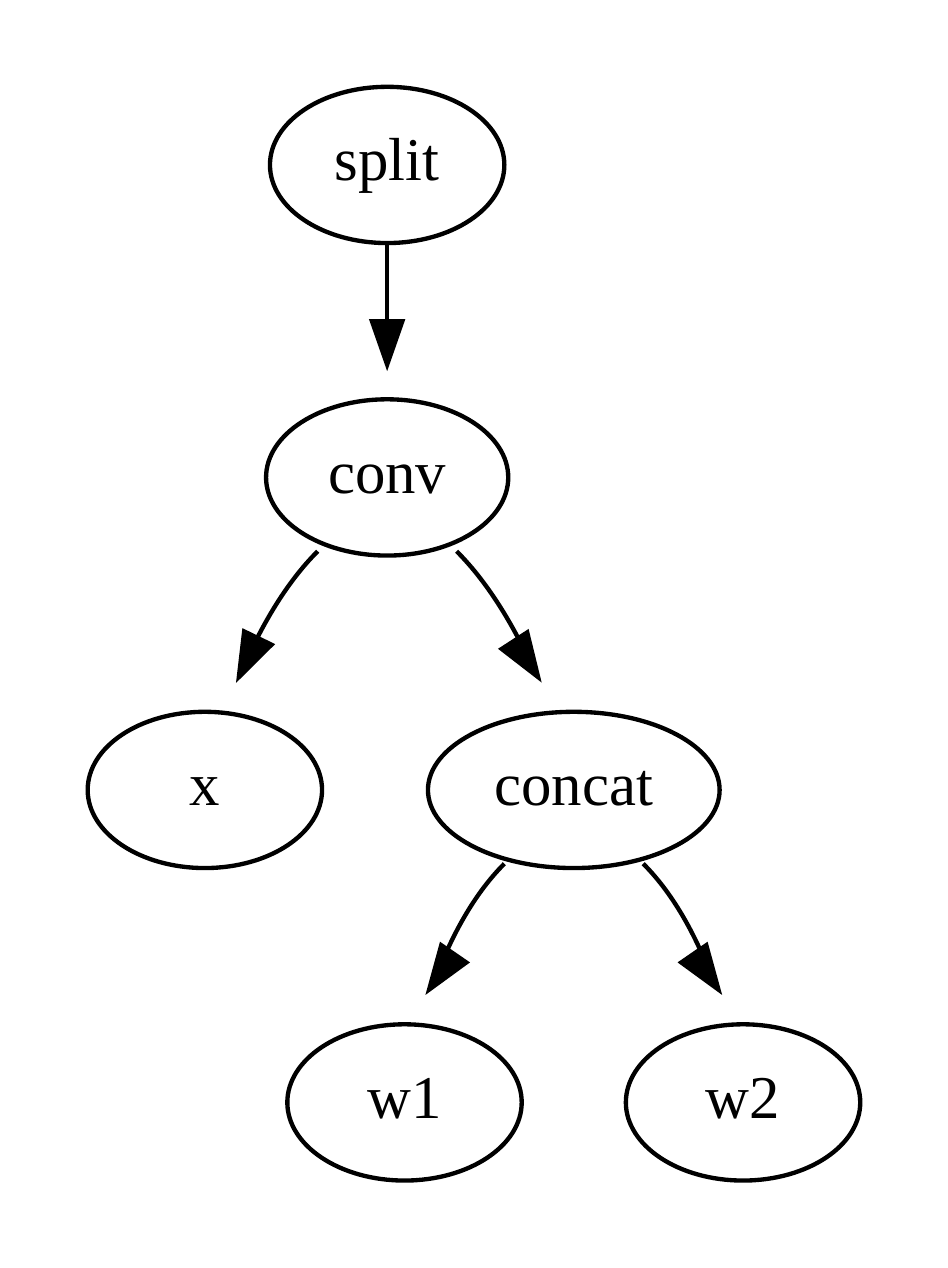}
    \hfill
    \caption{Example pattern used in NasNet-A and Inception-v3. 
    The optimized graphs also contain this pattern generalizing to more than 2 \texttt{conv}s sharing a common input node.}
    \label{fig:incept-rewrite}
\end{figure}

\begin{figure*}[t]
    \centering
    \includegraphics[width=0.49\linewidth]{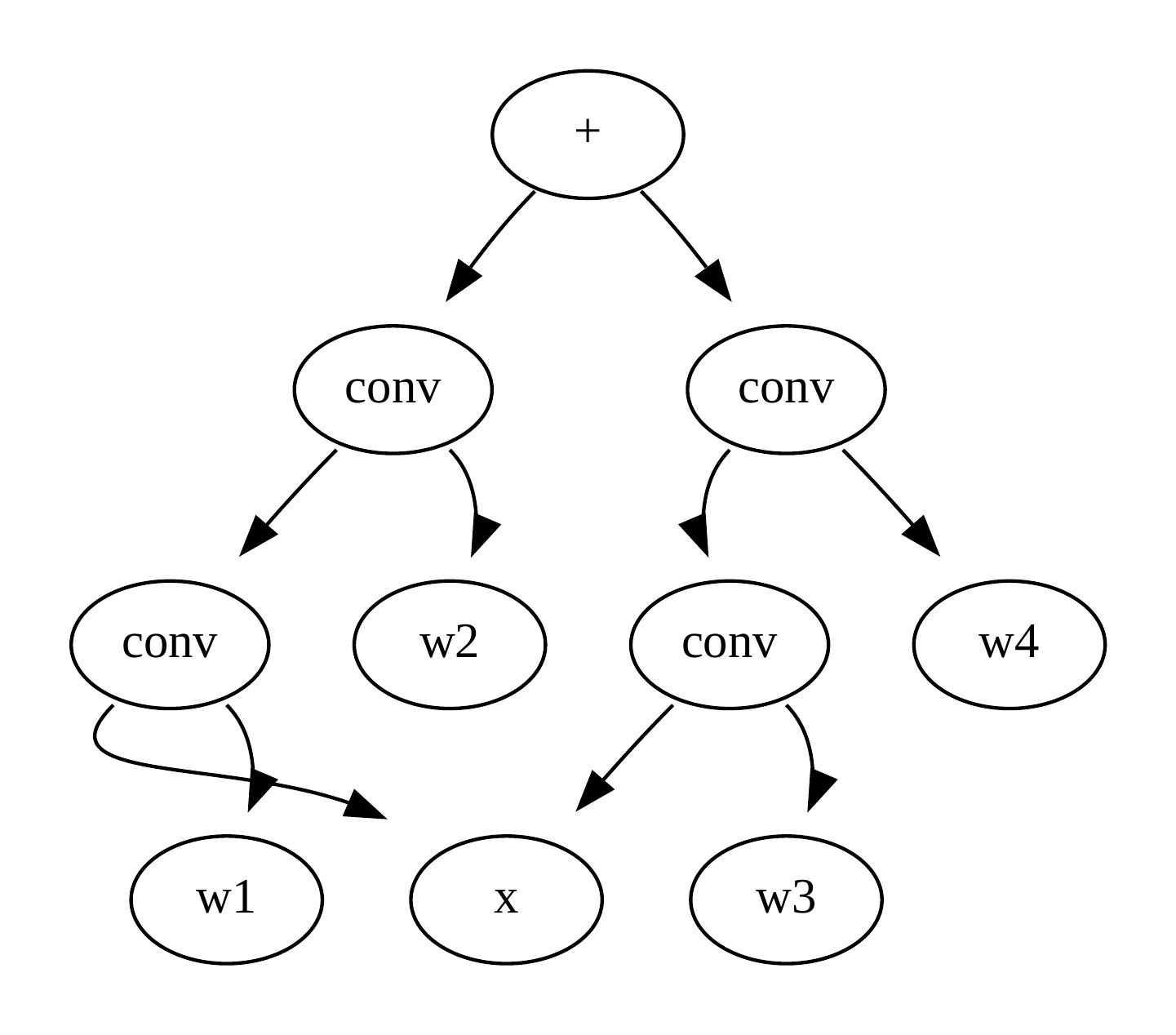}
    \includegraphics[width=0.49\linewidth]{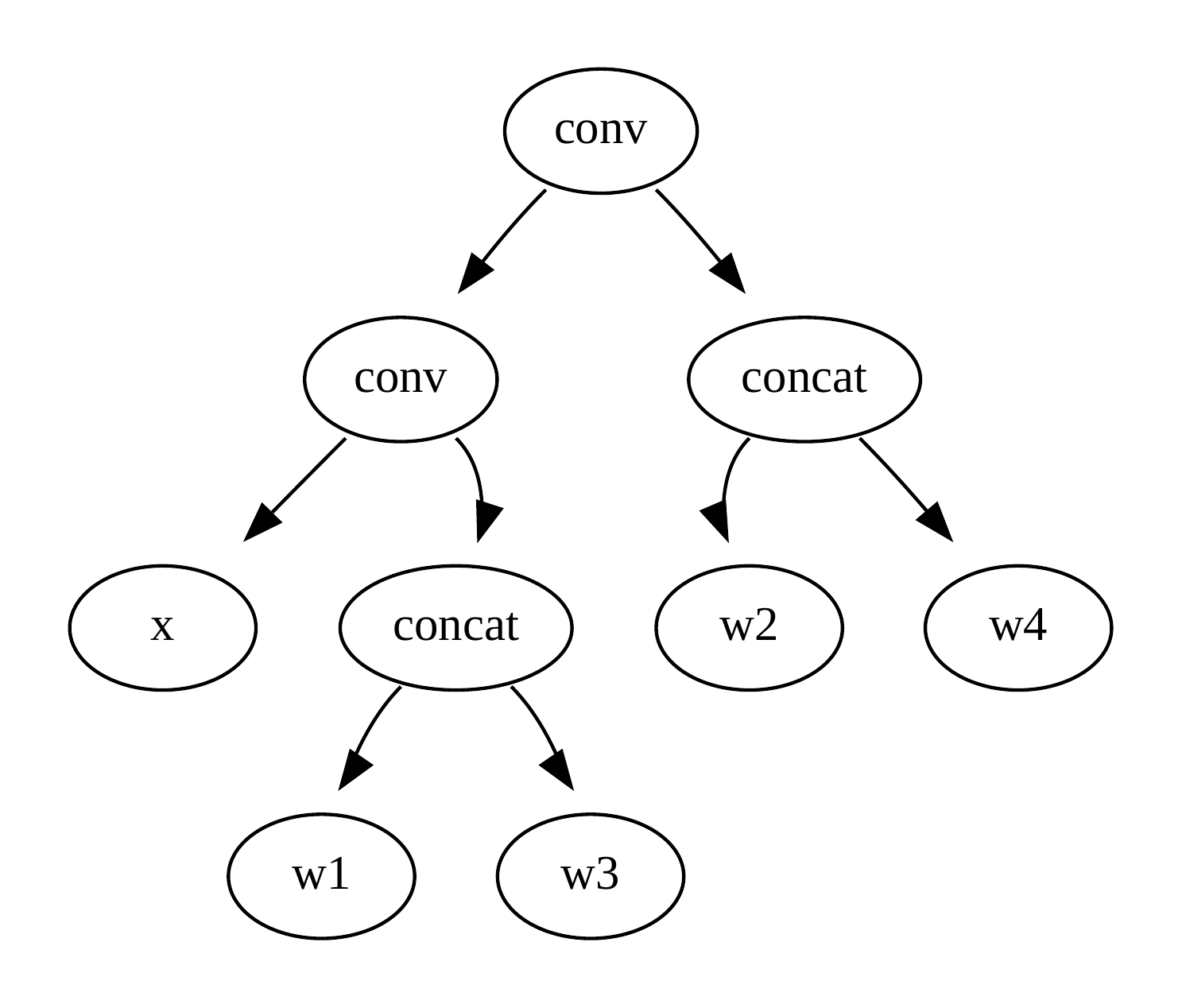}
    \caption{An example pattern of rewrite useful for NasNet-A. 
    Left: pattern in the original graph. 
    Right: pattern in the optimized graph by \ourname{}. 
    We only show the core operator nodes for clarity.
    Here, each $w_i$ is a convolution weight kernel. 
    The dimensions are ordered by \texttt{(out\_channels, in\_channels, height, width)}.
    \texttt{concat(w1, w3)} is over \texttt{axis=0} (output channels), and \texttt{concat(w2, w4)} is over \texttt{axis=1} (input channels).
    Since the two \texttt{concat} operators only involve weight nodes as inputs, they can be pre-computed in inference time. 
    Therefore, this rewrite pattern effectively convert four convolutions into two.
    }
    \label{fig:NasNet-A-rewrite}
\end{figure*}

\begin{figure*}[t]
    \centering
    \includegraphics[width=0.49\linewidth]{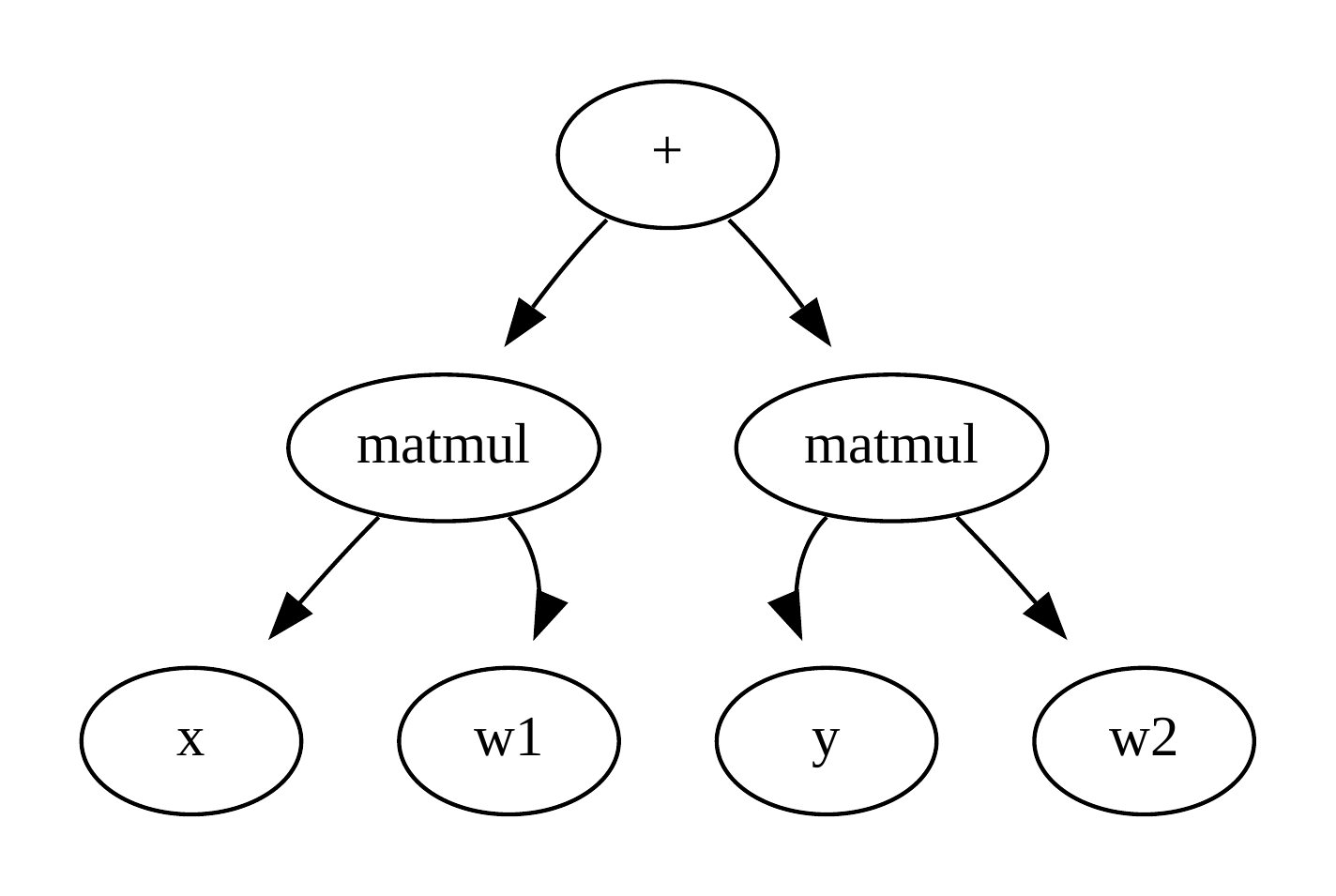}
    \includegraphics[width=0.49\linewidth]{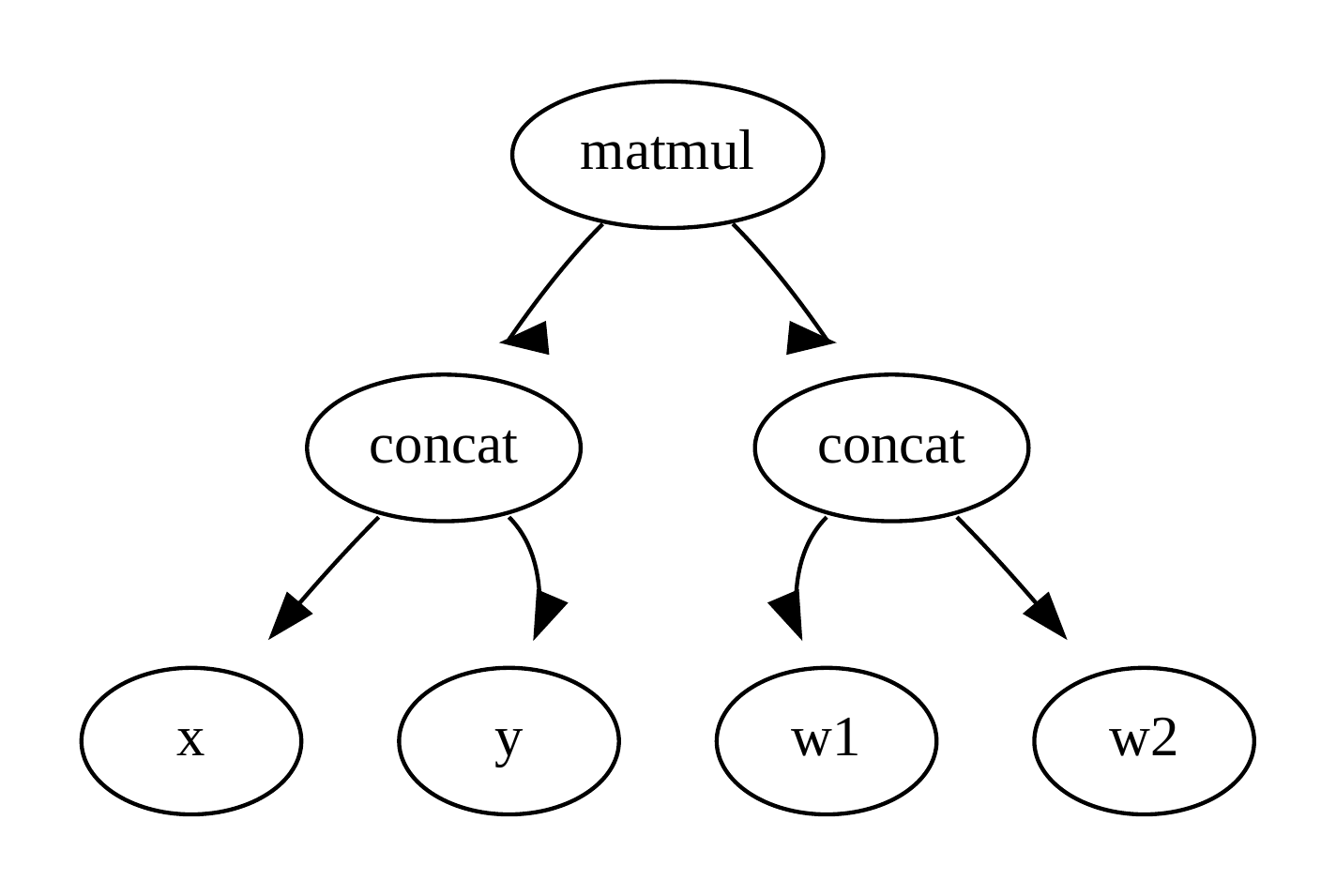}
    \caption{Example pattern used in NasRNN. 
    The optimized graphs also contain this pattern generalizing to more than 2 \texttt{matmul}s sharing a common input node.}
    \label{fig:NasRNN-rewrite}
\end{figure*}

\eat{
\begin{figure*}[h]
    \centering
    \includegraphics[width=0.4\linewidth]{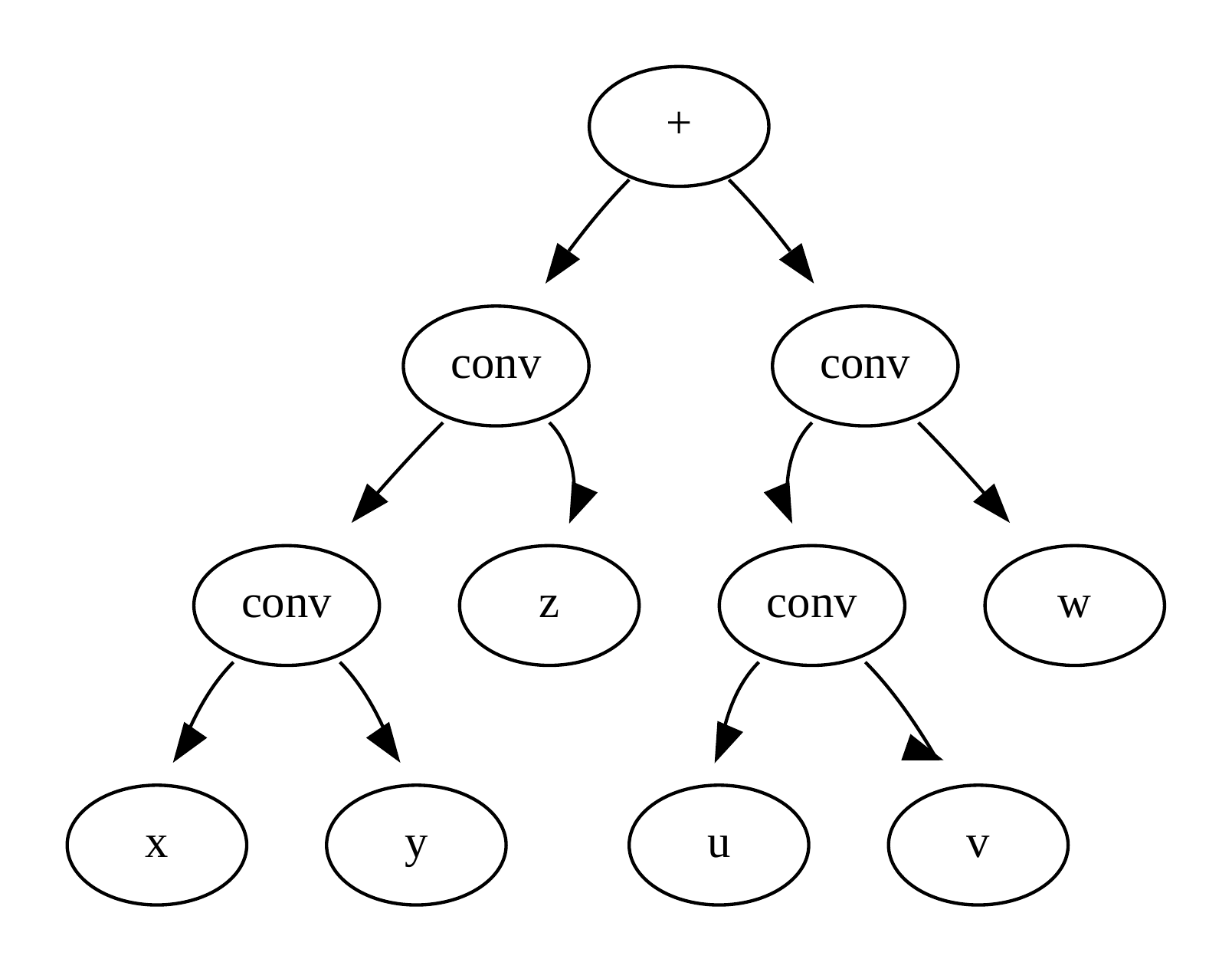}
    \includegraphics[width=0.4\linewidth]{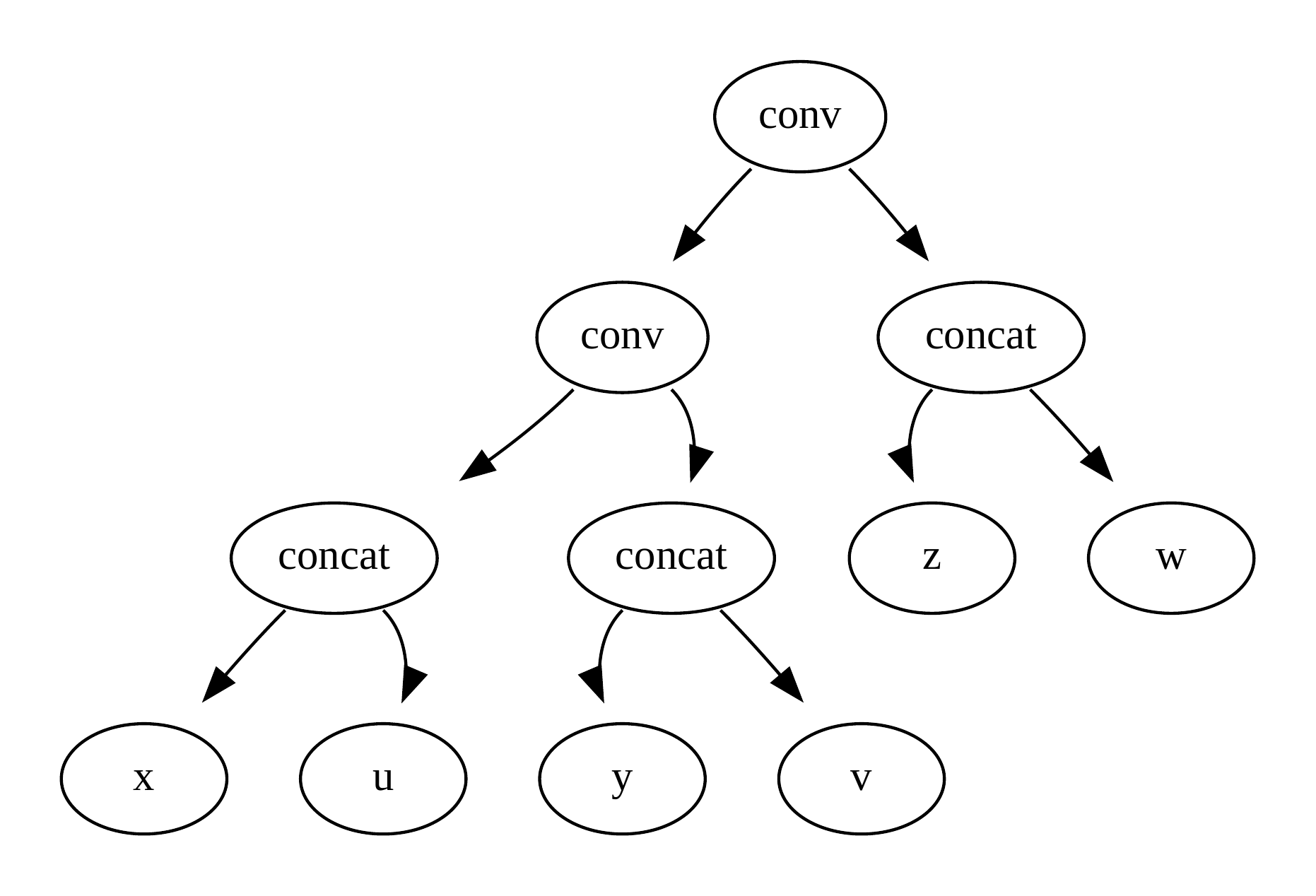}
    \caption{Caption}
    \label{fig:NasRNN-rewrite}
\end{figure*}
}

\end{document}